\pdfoutput=1
\documentclass[11pt]{article}
\usepackage[final]{acl}

\usepackage{times}
\usepackage{latexsym}
\usepackage[T1]{fontenc}
\usepackage[utf8]{inputenc}
\usepackage{microtype}
\usepackage{inconsolata}
\usepackage{graphicx}
\usepackage{subcaption}
\usepackage{booktabs}
\usepackage{tabularx}
\usepackage{hyperref}
\usepackage{amsmath}

\usepackage{longtable}
\usepackage{enumitem}
\usepackage{multirow}

\usepackage{annotates}
\usepackage{mparhack}
\providecommentcommand{rom}{blue}{Roman}
\providecommentcommand{jia}{red}{Jiahui}
\providecommentcommand{sea}{green!30!black}{Sean}
\usepackage{siunitx}
\usepackage{hyperref}

\title{Are Humans as Brittle as Large Language Models?} 

\author{Jiahui Li, Sean Papay \and Roman Klinger \\
    Fundamentals of Natural Language Processing, University of Bamberg, Germany\\
    \texttt{\{jiahui.li,sean.papay,roman.klinger\}@uni-bamberg.de}}

\begin{document}
\maketitle
\begin{abstract}
  The output of large language models (LLMs) is unstable, due both to
  non-determinism of the decoding process as well as to prompt
  brittleness. While the intrinsic non-determinism of LLM generation
  may mimic existing uncertainty in human annotations through
  distributional shifts in outputs, it is largely assumed, yet
  unexplored, that the prompt brittleness effect is unique to LLMs.
  This raises the question: do human annotators show similar
  sensitivity to prompt changes? If so, should prompt
  brittleness in LLMs be considered problematic? One may
  alternatively hypothesize that prompt brittleness correctly reflects
  human annotation variances. To fill this research gap, we
  systematically compare the effects of prompt modifications on LLMs
  and identical instruction modifications for human annotators,
  focusing on the question of whether humans are similarly sensitive
  to prompt perturbations. To study this, we prompt both humans and
  LLMs for a set of text classification tasks conditioned on prompt
  variations. Our findings indicate that both humans and LLMs exhibit
  increased brittleness in response to specific types of prompt
  modifications, particularly those involving the substitution of
  alternative label sets or label formats. However, the distribution
  of human judgments is less affected by typographical errors and
  reversed label order than that of LLMs.
\end{abstract}

\section{Introduction}
Large language models (LLMs) have shown impressive capabilities in
automatic data annotation tasks~\citep{tan-etal-2024-large}. However,
practical applications are often hindered by variability in model
outputs, leading to inconsistent
predictions~\citep{zhao-etal-2024-measuring,DBLP:journals/corr/abs-2405-01724}. This variability encompasses
both the inherent non-determinism in probabilistic
models~\citep{song-etal-2025-good,10.1145/3697010} as well as
\textit{prompt brittleness}, wherein minor changes in prompt phrasing
lead to significant differences in
outputs~\citep{lu-etal-2022-fantastically,tang-etal-2024-found}.

Of course, variability is also observed in human annotations.
Annotators’ individual experiences, sociodemographic backgrounds, and
moral values shape how they interpret and label content.
Moreover, the uncertainty in human annotation behavior may increase
with the inclusion of additional annotators, posing challenges for
explanation and identification~\citep{inherentdis}. Traditionally,
such disagreement among annotators has been treated as noise or
bias~\citep{milkowski-etal-2021-personal}, often resolved through
techniques like majority voting to produce a single ``gold''
label~\citep{sabou-etal-2014-corpus,anand-etal-2024-dont}.

Recent research suggests that variability in human annotations should
not be seen as a ``problem'' but as a reflection of diverse human
interpretations \citep{plank-2022-problem}. For example, in an
emotion labeling task, the text \textit{`the dog ran towards me.'} may
be labeled as joy or fear, depending on subjective preferences \citep{troiano-etal-2023-dimensional}.

As LLMs are trained to model human text, it is reasonable that their
output distributions should somehow reflect the variability present in
human annotations. However, while recent work has acknowledged the
importance of human label variation, little attention has been paid to
how this variation behaves across different prompt perturbations,
particularly whether changes in annotator instructions affect the
distribution of human responses analogously to LLM's prompt brittleness.

In this work, we draw a connection between human label variation and
LLM predictions. Specifically, we investigate whether prompt
brittleness -- a model’s sensitivity to small changes in task phrasing
-- is unique to LLMs, or if humans exhibit a comparable sensitivity
to instruction variations.
%which we refer to as \textit{human brittleness} in this paper.
%I reverted your change here, since this is the only point in the paper you refer to human brittleness.
We propose a systematic method that formalizes prompt perturbation and
investigates the distributional effects of such perturbations on both
LLMs and human annotators.  This method supports our exploration of
the following research questions:
\begin{description}[nosep]
    \item[RQ1:] How do distributional shifts of LLM outputs reveal prompt brittleness?
    \item[RQ2:] Are humans susceptible to prompt brittleness in ways comparable to LLMs?
    \item[RQ3:] Which types of prompt variants yield similar responses between humans and LLMs?
\end{description}

The experimental repository and results are available at
\url{https://www.uni-bamberg.de/en/nlproc/projects/inprompt/}.

\section{Related Work} \label{sec:related}

In the following, we discuss related work on variability in LLM outputs and human annotations.

\subsection{Variability in LLMs} \label{sec:brittle}
We focus on two factors for variability in LLMs: prompt brittleness and model non-determinism.
\subsubsection{Prompt Brittleness}  
LLMs exhibit high sensitivity to variations in a prompt, even when
changes are
minimal~\citep{LMlearner,jiang-etal-2020-know,gao-etal-2021-making}. This
phenomenon is commonly referred to as prompt brittleness. Prior
studies have examined a variety of prompt perturbations that can lead
to substantial changes in model outputs.

One prominent issue is position bias, where LLMs exhibit preferences
for labels appearing in specific positions within a prompt. Altering
the order of answer options can substantially affect model
behavior~\citep{lu-etal-2022-fantastically,qiang-etal-2024-prompt,wang-etal-2024-large-language-models-fair,wang-etal-2024-answer-c}. Similarly,
the position of the set of possible answers as a whole affects the
output \citep{zheng2024large}.

Lexical perturbations such as typos or synonym substitutions also
affect the output. While some findings suggest that LLMs are
relatively robust to typographical
errors~\citep{movva-etal-2024-annotation, wang2024look}, other work
highlights brittleness in response to paraphrased instructions,
including variations in verbs, nouns, and
conjunctions~\citep{mizrahi-etal-2024-state}.
\citet{ceron-etal-2024-beyond}
explore the effect of instruction styles, comparing personal versus
impersonal phrasing.

Various approaches exist to mitigate the effect of prompt variations,
such as guiding the inference process
\citep{yugeswardeenoo-etal-2024-question,lampinen-etal-2022-language}
or by requesting explanations
\citep{Xprompt,mishra-etal-2022-cross}. Similarly, eliciting
verbalized confidence estimates can reduce calibration errors in model
outputs~\citep{tian-etal-2023-just}.

In this work, we will propose methods of prompt perturbations which build upon the
theoretical foundations laid out in these works.

\subsubsection{LLM Non-determinism}

LLMs not only yield inconsistent outputs due to prompt brittleness,
but also exhibit inherent non-determinism. Notably, even with a
temperature of zero, variability persists when invoking LLM APIs such
as ChatGPT~\citep{10.1145/3697010}. This stochastic behavior poses
challenges for reproducibility in downstream applications.

\citet{song-etal-2025-good} investigate the relationship between model
size and output stability, finding that smaller LLMs may yield more
consistent outputs when comparing the greedy decoding with different
sampling strategies. To address non-determinism in practical
(non-greedy) sampling scenarios, where LLMs produce a distribution of
outputs for the same prompt, \citet{hayes-etal-2025-measuring} propose
probabilistic discoverable extraction. 

In this work, we employ non-deterministic sampling-based decoding
strategies for LLMs, treating variability not as noise but as an
intrinsic property of LLM output, to be analyzed on the same footing
as human response variations.

\subsection{Variability in Humans}
In the context of human annotation, variations between annotators are
commonly treated as noise, often attributed to annotation errors
\citep{zhang-de-marneffe-2021-identifying}.  However,
\citet{plank-2022-problem} argues that variability in human
annotations can also arise from factors such as subjectivity,
perspectivism, annotator disagreement, and the existence of multiple
plausible views.  Conversely, in survey responses, inter-participant
variability is expected and desired in order to reflect the diversity
of the surveyed population.  However, such responses may also be
sensitive to how questions are asked \citep{survey-review}, exhibiting
similarities to the prompt-brittleness phenomenon observed for LLMs.

\subsubsection{Subjectivity \& Annotator Disagreement}
Subjective tasks, such as hate speech detection and emotion
classification, are inherently influenced by the value systems,
cognitive frameworks, and personal experiences of
annotators~\citep{sap-etal-2019-risk,truthlie,milkowski-etal-2021-personal}. 
Recent work has introduced this concept in humans to LLMs, e.g., with
socio-demographic
prompting~\citep{2025demo,dayanik-etal-2022-analysis}. Our approach
considers the distribution of predictions by prompting LLMs a large
number of times without anchoring responses to predefined demographic
categories.

While subjectivity and perspectivism highlight the inherent
variability in human interpretation, annotator disagreement also
points to varying annotations which might not be of any
benefit. Disagreement is therefore also observed in tasks considered
objective, demanding a unified consensus for both model training and
evaluation~\citep{disagree}.

For example, \citet{plank-etal-2014-linguistically} demonstrate that
disagreements in part-of-speech tagging stem from linguistically
debatable cases rather than annotator mistakes. Similarly,
\citet{webber-joshi-2012-discourse} analyze the challenges of human
disagreement in discourse properties. 
Semantic annotation tasks are also prone to disagreement.
\citet{sommerauer-etal-2020-describe} evaluate such cases using
multiple quality metrics. De Marneffe (\citeyear{didithappen}) argue that fact
judgments should be treated as distributions. 

In contrast to prior work that explores why annotators disagree, our
study focuses on how both humans and LLMs vary in a distribution of
their responses, considering the above cases as a whole.

\subsubsection{Survey Response Bias}

In social science studies, human responses often exhibit biases
influenced by different survey designs, such as variations in wording,
synonyms, and question formats used in market research
\citep{brace2008questionnaire,responseAlt}.  Also, the order of answer
options in questionnaires can give rise to acquiescence and recency
effects \citep{responseOrd}.

\citet {tjuatja-etal-2024-llms} present the only study we are aware of
which studies the relation between LLMs and human variations.
They demonstrate that
commercial LLMs exhibit sensitivity to prompt biases that elicit
minimal change in human responses. However, their exploration of
prompt modifications primarily relies on survey-style prompt variants
and does not study human annotation variability. In contrast, our work
systematically generalizes a broader and more diverse set of
prompt reformulations, aiming to reflect the correlation of
variability between LLMs and human annotation behavior.

\begin{table*}[t]
\centering
\small
\begin{tabularx}{\textwidth}{p{1cm}p{1.8cm}p{3cm}p{3.2cm}X}
\toprule
\textbf{Name} & \textbf{Type } & \textbf{Description} & \textbf{Variant} & \textbf{Example}\\
\midrule
Base & Base Prompt & A single base prompt to apply modifications. & --- & How do you rate ... given one of the labels `a', `b', `c', `d'?  \\
\midrule
Imper & Imperative expression & Reframes the prompt as an imperative. & pls: inserts the word `please' for politeness. & \textbf{Please} rate ... given one of the labels `a', `b', `c', `d'\textbf{.} \\
\midrule
LabelOrd & Label order & Alters the internal order of provided labels within the prompt. & 1. rev: reverse the labels order.   \newline 2. shuff: shuffle the labels at random. & 1. How do you rate ... given one of the labels \textbf{`d', `c', `b', `a'}? \newline 2. How do you rate ... given one of the labels \textbf{`a', `d', `b', `c'}? \\
\midrule
LabelPos & Label \newline position & Specifies the location of the provided label set as a whole. & 1. start: presents at the start of the prompt. \newline 2. end: presents at the end of the prompt. & 1. \textbf{Given one of the labels} `a', `b', `c', `d', how do you rate ...?\newline 2. How do you rate ... given one of the labels `a', `b', `c', `d'? \\
\midrule
Syns & Synonym substitution & Replaces lexical items with synonyms, applied to specific part-of-speech categories. & 1. verb: verbs.\newline 2. noun: nouns excluding labels.\newline 3. prep: preposition.\newline 4. co: coordinating conjunction. & 1. How do you \textbf{evaluate} ... given one of the labels `a', `b', `c', `d'?\newline 
  2. How do you rate ... given one of the \textbf{scales} `a', `b', `c', `d'?\newline
 3. How do you rate ... \textbf{according to} one of the labels `a', `b', `c', `d'?\newline 
 4. How do you rate ... given one of the labels `a', `b', `c', \textbf{or} `d'? \\
\midrule
Typo & Typographical errors & Introduces minor typographical errors into the prompt wording. & 1. task: typos in the content of task description. 2. label: typos in the provided labels. & 1. How do you \textbf{raet} ... given one of the labels `a', `b', `c', `d'?\newline 2. How do you rate ... given one of the labels `a', \textbf{`$\beta$'}, `c', `d'? \\
\midrule
Cap & Capitalization & Capitalizes the words in the prompt. & 1. task: key words of task content.\newline 2. label: all the provided labels. & 1. How do you rate ... given one of the \textbf{LABELS} `a', `b', `c', `d'?\newline
 2. How do you rate ... given one of the labels \textbf{`A', `B', `C', `D'}? \\
\midrule
PM & Punctuation mark & Operates on punctuation marks in the prompt. & 1. remove: removes existing punctuation.\newline 2. add: adds punctuation. \newline 3. replace: replaces punctuation with equivalent alternatives. & 1. How do you rate ... given one of the labels \textbf{a, b, c, d}?\newline 2. How do you rate ... given one of the labels\textbf{:} `a', `b', `c', `d'?\newline
 3. How do you rate ... given one of the labels `a'\textbf{;} `b'\textbf{;} `c'\textbf{;} `d'?\\
\bottomrule
\end{tabularx}
\caption{Prompt modification methods in the neutral
  category. Modifications are bold in the text for each example. Our hypothesis for these prompt variations is that human's show lower
  variation than LLMs.}
\label{tab:N}
\end{table*}

\section{Prompt Perturbation Methods} \label{sec:method}
To study the relation between human and model ``brittleness'', we
employ a systematic method for generating variations from a single
base prompt.  We hypothesize that prompt perturbations may be
categorized according to their potential influence on human annotation
decisions, which may reflect the variability of LLMs.  Thus, in order
to obtain a sufficient diversity of prompt variations, we explicitly
construct prompt modifications of two categories: \textit{neutral}
prompt modifications, which we hypothesize should not significantly
affect human responses, and \textit{sensitive} prompt modifications,
which we hypothesize should.

Within each category, we specify a number of aspects of the base
prompt which could be mutated, and, for each aspect, define one or
more concrete prompt variations.  Tables~\ref{tab:N} and \ref{tab:S}
summarize the modification methods employed in both categories.  For
example, in the neutral category, one aspect, \textit{LabelOrd},
reflects the internal order of the provided labels.  For this aspect,
we specify two concrete variants: \textit{rev}, which reverses the
base prompt's label order (mutating from ascending to descending
order), and \textit{shuff}, which permutes the labels randomly.

\textit{Sensitive} changes might include the introduction of emotionally connotated
wording in the prompt (referred to as \textit{Emo}).  For this aspect,
appending the sentence ``Respond quickly!'' to the prompt corresponds to
the variant \textit{fast}.  Appendix~\ref{app:prompts} provides the
full list of evaluated prompts in our experiments, developed by these
methods.

Importantly, prompt variations can modify both model input and the
label set, meaning model outputs and human annotations must be
interpreted with respect to the prompt variation.

\begin{table*}[t]
\centering
\small
\begin{tabularx}{\textwidth}{p{0.55cm}p{1.25cm}p{2.1cm}p{3.55cm}X}
\toprule
\textbf{Name} & \textbf{Type } & \textbf{Description} & \textbf{Variant} & \textbf{Example}\\
\midrule
AltLab & Alternative labels & Modifies the label set using synonymous or alternative mapping formats like Likert scales. & 1. gran: changes the granularity of the label set.\newline 2. int: changes the intensity of the label set.\newline 3. keep: synonyms of labels. & 1. How do you rate ... given one of the labels `a', \textbf{`a+'}, `b', \textbf{`b+'}, `c', `d'? \newline 2. How do you rate ... given one of the labels `a', `b', `\textbf{d}', `\textbf{e}'? \newline 3. How do you rate ... given one of the labels \textbf{`$a'$', `$b'$', `$c'$', `$d'$'}? \\
\midrule
Def & Definition insertion & Adds definitions of the task content to the prompt. & 1. task: explains the meaning of the task. \newline 
2. label: explains the meanings of labels. \newline
3. both: explains the meanings of both 1. \& 2. & 1. How do you rate ... given one of the labels `a', `b', `c', `d'? \textbf{... means...}\newline  2. How do you rate ... given one of the labels? \textbf{`a':..., `b':..., `c':..., `d':...}  \\
\midrule
Conf & Confidence statement & Request a confidence score with the answer. &---& How do you rate ... given one of the labels `a', `b', `c', `d'? \textbf{Provide your answer alongside a confidence score.} \\
\midrule
Exp & Explanation request & Asks for a justification. & --- & How do you rate ... given one of the labels `a', `b', `c', `d'? \textbf{Provide your answer with a justification.} \\
\midrule
Emo & Emotional wording & Adds emotionally charged content to the prompt.  & 1. trust: requires trust of the answer. \newline 2. warn: content warning. \newline 3. care: requires care when making the answer. \newline 4. fast: requires completing the task fast.\newline & 1. How do you rate ... given one of the labels `a', `b', `c', `d'? \textbf{Trust your answer!} \newline 
 2. How do you rate ... given one of the labels `a', `b', `c', `d'? \textbf{The text may contain offensive words.}\newline
 3. How do you rate ... given one of the labels `a', `b', `c', `d'? \textbf{Be careful with your answer.}\newline
 4. How do you rate ... given one of the labels `a', `b', `c', `d'? \textbf{Respond fast!} \\
\bottomrule
\end{tabularx}
\caption{Prompt modification methods in the \textit{sensitive}
  category. Modifications are bold in the text for each example. We
  assume that humans are more affected by these variations than by
  those in the neutral category.}
\label{tab:S}
\end{table*}

\section{Experimental Settings} \label{sec:exp}
Our experiments involve collection of data from two sources: LLM predictions and human annotations.
In order to compare LLM prompt brittleness to any potential instruction-conditioned variability in humans,
we adopt an experimental methodology which maintains parallelism between these two data sources.
Thus, we will first discuss our datasets and tasks, which are common to both, before detailing
our specific settings for obtaining LLM predictions and human annotations.

\paragraph{Datasets and Tasks.}
For both humans and LLMs, we consider four tasks from three distinct
English-language datasets: (1) offensiveness rating and (2) politeness
rating from the \textsc{Popquorn}
dataset~\citep{pei-jurgens-2023-annotator}, (3) irony detection from
the \textsc{EPIC} dataset~\citep{frenda-etal-2023-epic}, and (4)
emotion classification from the \textsc{crowd-enVent}
dataset~\citep{troiano-etal-2023-dimensional}.  These tasks were
selected as all are known to involve some level of annotator
subjectivity.
For each dataset, we construct a base prompt, which clearly and
succinctly requests an annotation for a provided instance according to
the task's label set. We then construct a set of prompt variations,
as described in Section~\ref{sec:method}. A full list of all prompt
variations thus obtained is detailed in Appendix~\ref{app:prompts}.

\begin{table}[t]
\centering
\begin{tabular}{lcr}
\toprule
\textbf{Task} & \textbf{Inst.} & \textbf{AnnNum} \\
\midrule
Offensiveness & 1477 & 8.7\\
Politeness & 3704 & 6.7\\
Irony & 2987 & 5.0\\
Emotion & 1161 & 5.0\\
\bottomrule
\end{tabular}
\caption{Statistics of the evaluated datasets. \textit{Inst.}: Number of retained instances; \textit{AnnNum}: Average number of annotators per instance.}
\label{tab:dataset_stats}
\end{table}

\paragraph{LLMs.}
We select five LLMs capable of local deployment for evaluation:
LLaMA-3.1-8B and LLaMA-3.3-70B~\citep{llama3},
Mixtral-8x7B~\citep{mixtral}, Falcon3-7B~\citep{Falcon3}, and
Mistral-7B~\citep{mistral7b}.\footnote{\url{https://huggingface.co/}:
  \texttt{tiiuae/Falcon3-7B-Instruct},
  \texttt{mistralai/Mixtral-8x7B-Instruct-v0.1},
  \texttt{mistralai/Mistral-7B-Instruct-v0.3},
  \texttt{meta-llama/Llama-3.3-70B-Instruct},
  \texttt{meta-llama/Llama-3.1-8B-Instruct}.}

All models are accessed and executed using the HuggingFace library
ecosystem. In order to investigate the distribution of LLM outputs,
we enable stochastic decoding via sampling during generation.
Specifically, the decoding configuration was selected to match typical
deployments, with \texttt{top\_p=0.9}, \texttt{temperature=0.6}, and
\texttt{top\_k=50}. These settings correspond to the default
configuration used for the LLaMA-3.1-8B and LLaMA-3.3-70B models in
the HuggingFace transformers library. The experiments are run on
Nvidia A40 and Nvidia A100 GPUs, with a total estimation of 13,500 GPU
hours for our study.

Model predictions are elicited in a zero-shot fashion: For each
instance, the model is presented with one prompt variation, followed
by the instance text. The generated
text is then parsed by attempting to extract one of the requested
labels, with unparsable responses discarded. Across all tasks, for
each (prompt variation, instance, LLM) triple, we elicit 100 responses
in this manner. We discard instances for which we obtain less than
100 total valid predictions after 500 attempts. This leaves us with a total of
9329 retained instances across all tasks, for which the statistics
are shown in Table~\ref{tab:dataset_stats}.

\begin{figure*}
  \centering

  \begin{subfigure}[t]{\textwidth}
    \includegraphics[width=\linewidth]{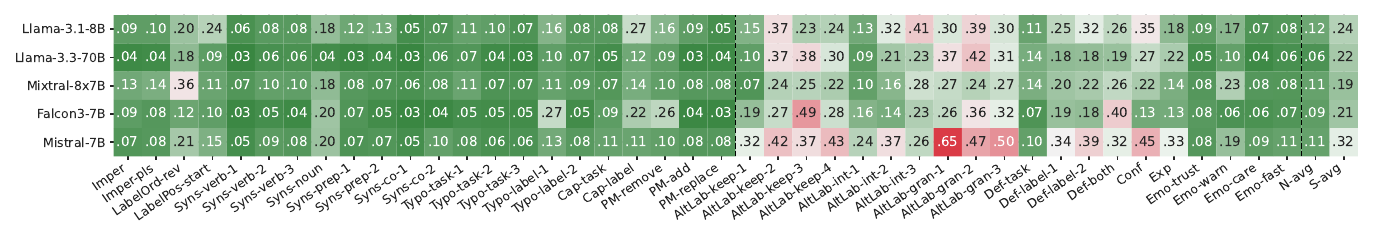}
    \caption{Offensiveness rating.}
    \label{fig:off-all}
  \end{subfigure}

  \begin{subfigure}[t]{\textwidth}
    \includegraphics[width=\linewidth]{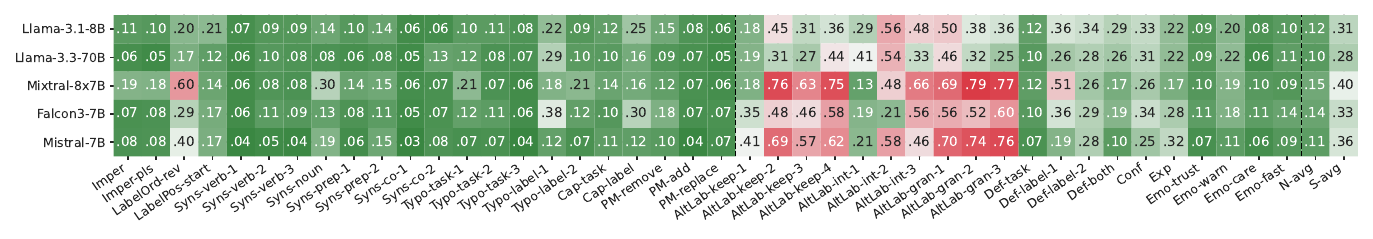}
    \caption{Politeness rating.}
    \label{fig:po-all}
  \end{subfigure}

  \begin{subfigure}[t]{\textwidth}
    \includegraphics[width=\linewidth]{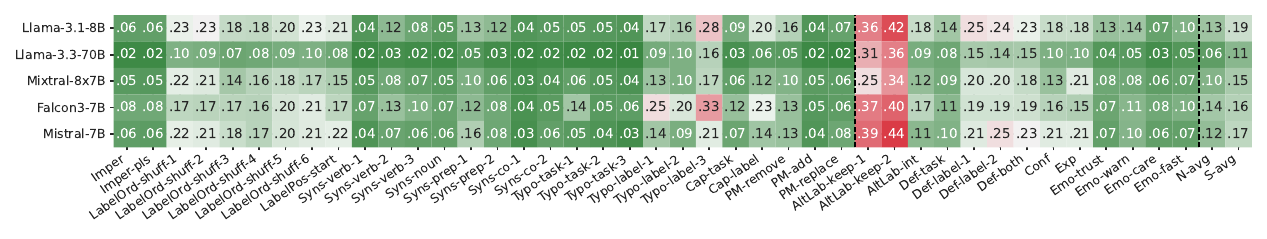}
    \caption{Emotion classification.}
    \label{fig:emo-all}
  \end{subfigure}

  \begin{subfigure}[t]{\textwidth}
    \includegraphics[width=\linewidth]{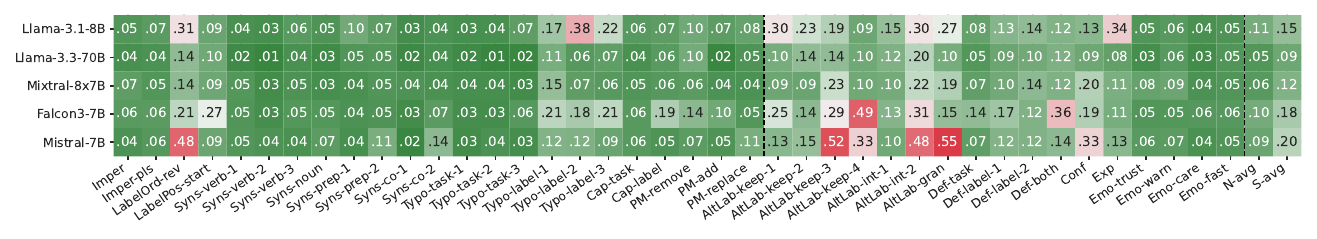}
    \caption{Irony detection.}
    \label{fig:iro-all}
  \end{subfigure}

  \caption{
    Heatmaps showing distance scores between the distributions of LLM predictions with prompt variants and with the base prompt for four evaluated tasks in Table~\ref{tab:dataset_stats}.
    Distance scores are calculated using Jensen--Shannon divergence~\citep{dagan-etal-1997-similarity}. The x-axis represents different types of prompt modifications defined in Section~\ref{sec:method}, with numeric indices representing multiple instances of the same type-variant pair. The y-axis lists the five LLMs evaluated in the study. Black dashed lines divide each heatmap into three parts: prompt modifications belonging to the neutral class (left), those from the sensitive class (middle), and the average distance scores for neutral and sensitive class modifications respectively (right). All prompts evaluated are provided in Appendix~\ref{app:prompts}.
  }
  \label{fig:all-heatmaps}
\end{figure*}

\begin{figure*}
  \centering
\begin{subfigure}[t]{\columnwidth}
   
  \includegraphics[width=0.7\columnwidth]{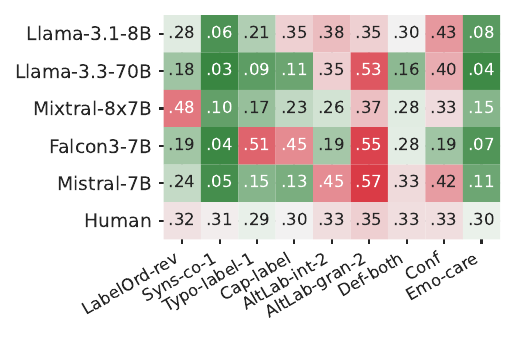}
  \caption{Offensiveness rating.}
  \label{fig:off-human}
\end{subfigure}
% \vspace{0.3cm}
\begin{subfigure}[t]{\columnwidth}
\centering
  \includegraphics[width=0.7\columnwidth]{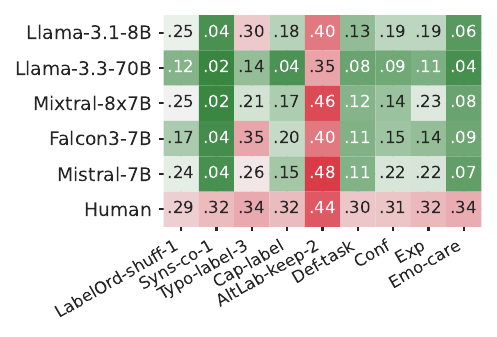}
  \caption{Emotion classification.}
  \label{fig:emo-human}
\end{subfigure}
\caption{Distance scores between the distributions of responses with prompt variants and the base prompt. The evaluated tasks are offensiveness rating (a) and emotion classification (b). The x-axis refers to the prompts with their modification types, and the y-axis refers to five LLMs and human samples. Distance scores are measured by Jensen--Shannon divergence~\citep{dagan-etal-1997-similarity}. Evaluated prompts can be found in Table~\ref{tab:prompt-off} (a) and Table~\ref{tab:prompt-emo} (b).}
\label{fig:human-study}
\end{figure*}

\paragraph{Human Study.}
Due to the high cost of human annotation, we collect a subset of the judgments from our LLM study for human evaluation, focusing on two tasks: offensiveness rating and emotion classification. For each task, we sample approximately 10\% of
instances from the original \textsc{Popquorn} and
\textsc{crowd-enVent} datasets, selecting an equal number of instances
for each gold-standard label. We use ten prompts per task for the
human study, selected so as to maximize the diversity of prompts
tested. In addition to the base prompt, we select six variants which
caused the most, medium and least extreme changes in LLM predictions from both
the \textit{sensitive} and \textit{neutral} categories, as measured by Jensen--Shannon
divergence (see Section~\ref{sec:analysis}). The other three variants are selected from the remaining types which present notable changes in instructing humans. Each prompt-instance
pair is annotated by ten independent annotators recruited via the
Prolific platform.\footnote{\url{https://www.prolific.com/}} We employ
only annotators based in the United States, with U.S.\ listed as their
country of residence, country of birth, and nationality. All
annotators are at least 18 years of age and demonstrated a high level
of English proficiency, with English specified as their primary,
first, and fluent language. We restrict the annotator approval rating
to greater than 99\%. Each annotator is presented with 22 instances
including two attention checks, each preceded by a prompt variation
acting as instructions. To mitigate potential biases
arising from prior exposure to the same task, each annotator is
assigned only one prompt variant per task. Each annotator is paid \pounds9 per hour as suggested by Prolific. The median completion time for each survey is approximately 7–8 minutes, except for the \textit{Exp} prompt variant, which requires about 16 minutes. The whole experiment costs \pounds2,400, including 33.3\% Prolific service fee.

\section{Analysis}
\label{sec:analysis}

In this section, we present the results and analysis by answering
three main research questions.

For both human annotators and LLMs, we quantify the distributional
differences between predictions elicited by each prompt variant and
those elicited by the base prompt. A full list of evaluated prompts
is provided in Appendix~\ref{app:prompts}. For all the tasks, we
employ the Jensen--Shannon
divergence~\citep{61115,dagan-etal-1997-similarity} as a measure of
distributional dissimilarity to display the human or LLM brittleness to prompt changes.

\subsection{RQ1: How do distributional shifts of LLM outputs reveal prompt brittleness?}
Figure~\ref{fig:all-heatmaps} illustrates the distributional distances
between various prompt variants and the corresponding base prompt
across the four evaluated tasks. Across all tasks and five LLMs, we
find that prompt changes categorized as \textit{neutral} generally
lead to smaller distributional shifts in model outputs compared to
those categorized as \textit{sensitive}. Interestingly, we observe that changes involving alternative label formulations, either semantically or in substitute mapping formats, exert the most important influence on model output
distributions. Modifications including adding definitions and
requiring justifications or a confidence score, which belong to the \textit{sensitive} category, exhibit a moderate effect.

While these results broadly align with our hypothesis on the
categorization of prompt types from the human-grounded aspect, some
exceptions are noteworthy. Within the \textit{neutral} category,
changes to the internal order of labels (\textit{LabelOrd}) induce
substantial distributional shifts compared to changes to other prompt
aspects.  Within the \textit{sensitive} category, introducing
emotional language results in a relatively weaker impact than other
modifications. Seemingly subtle changes, such as typos and
capitalization of labels, yield measurable distributional shifts,
highlighting LLM sensitivity to some surface variations, especially in
labels.

Although these findings provide strong evidence of prompt brittleness,
we observe that the LLaMA-3.3-70B model demonstrates greater
robustness to prompt variations across all tasks, as indicated by its
relatively lower average divergence scores. The Mixtral-8x7B model
ranks second in robustness, except in the politeness rating task.
This suggests that larger models tend to be more consistent to prompt
perturbations than smaller ones.

To assess whether different LLMs exhibit similar patterns of
brittleness across prompt modifications, we compute Spearman’s rank
correlation coefficients~\citep{spearman04} between pairs of models,
based on the ranking of divergence scores induced by each prompt
variant. Heatmaps of pairwise correlations are provided in
Appendix~\ref{app:spearmanr}. For the emotion classification task,
the models show high agreement in their ranking of prompt
sensitivities, indicating a shared trend of distributional shifts.
Although other tasks display lower rank correlations, the coefficients
remain positive, suggesting a generally similar trend to different
prompt modifications across models.

\begin{figure*}[t]
\centering

\begin{subfigure}[t]{\textwidth}
  \caption{Offensiveness rating.} \label{fig:off-matrix}
  \includegraphics[width=0.98\textwidth]{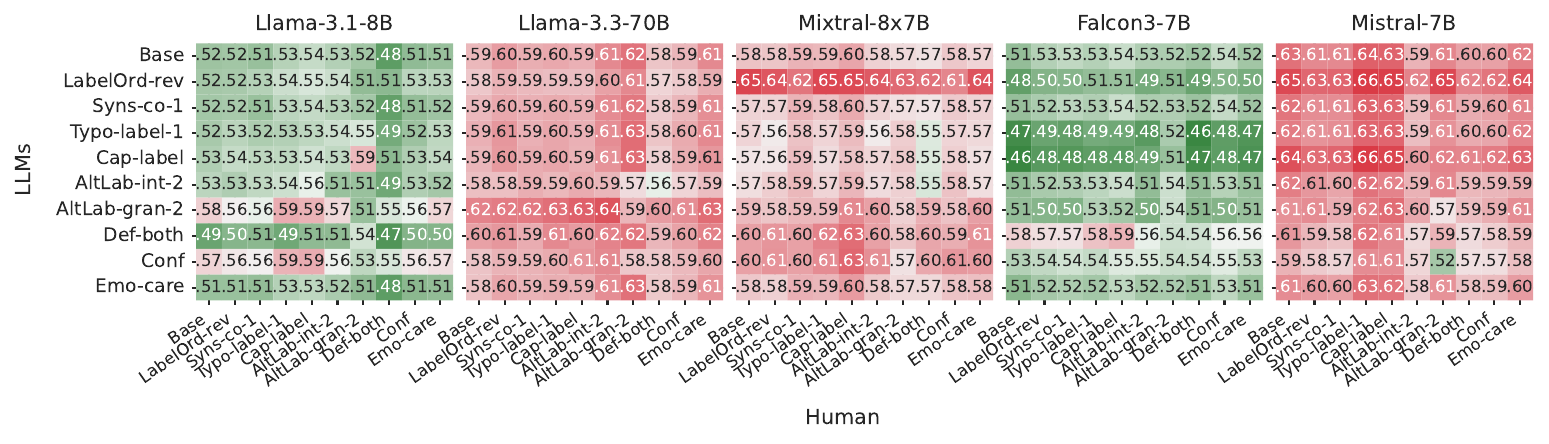}
  \end{subfigure}
  
  \begin{subfigure}[t]{\textwidth}
\centering
  \caption{Emotion classification.}
  \label{fig:emo-matrix}
  \includegraphics[width=0.98\textwidth]{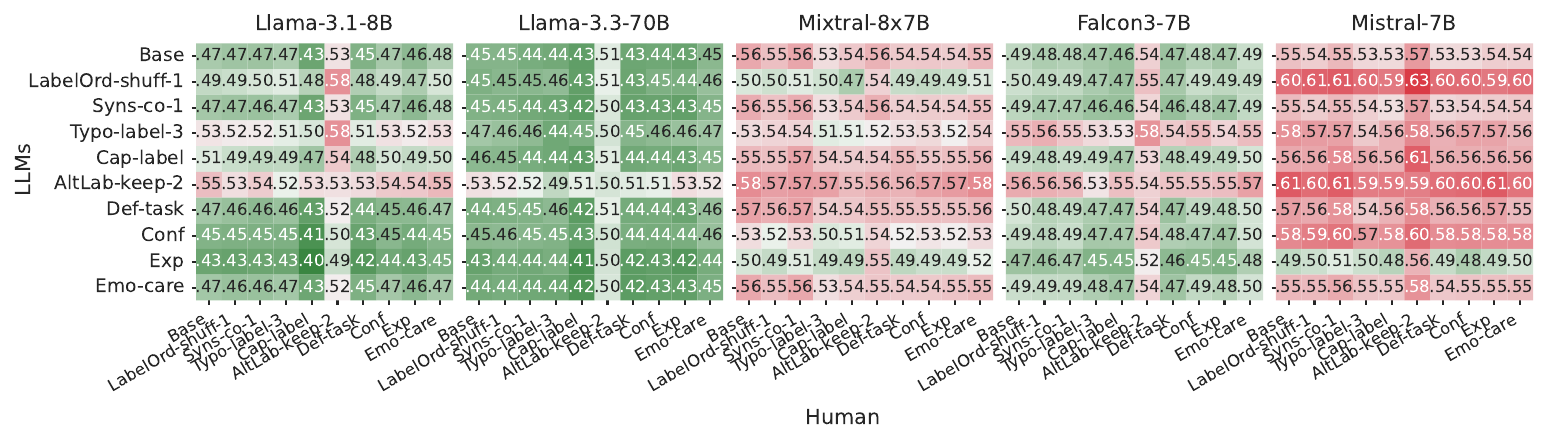}
  
\end{subfigure}
\caption{Distributional distance between LLM-generated and human annotations across five LLMs. The evaluated tasks are (a) offensiveness rating and (b) emotion classification. The x-axis and y-axis denote the prompt variants for human and LLMs respectively. The model name is displayed on the top of each subfigure. Evaluated prompts can be found in Table~\ref{tab:prompt-off} for (a) and Table~\ref{tab:prompt-emo} for (b).
  }
  \label{fig:matrix}
\end{figure*}

\subsection{RQ2: Are humans susceptible to prompt brittleness in ways comparable to LLMs?}

To investigate whether prompt brittleness also affects human annotators,
we turn to our human annotation study,
and compare the response distribution divergences across both LLMs and our human annotators.
These results are presented in Figure~\ref{fig:human-study}.

Overall, we find that human divergences are higher for the same prompt, but seem to be less dramatically
affected by the specifics of instruction variations than LLMs.
Across both tasks, humans have an average divergence of 0.32, compared to an average LLM divergence of 0.22,
but LLMs show a much larger standard deviation of 0.14, compared to 0.03 for humans.
This is largely consistent with a state of affairs where human empirical distributions show greater instability, likely due to their smaller sample
size, while showing less instruction-conditioned instability (i.e. brittleness) than LLMs.

However, while humans seem to be \textit{less} brittle than LLMs, we still observe
a clear brittleness effect in human annotations.
In order to quantify human sensitivity to instruction variations, we employ aggregated $\chi^2$-tests to compare each prompt variation
to the base prompt, pooled across all instances which we assume to be independent.
Table~\ref{tab:chi2} summaries our findings: at a $p<0.05$ level, we find statistically significant differences in human response distributions with
\textit{LabelOrd-rev}, \textit{AltLab-gran-2}, \textit{AltLab-int-2}, \textit{Def-both}, \textit{Conf}, and \textit{Def-both} for the offensiveness rating task, and with
\textit{Typo-label-3}, \textit{Emo-care}, \textit{AltLab-keep-2} for the emotion classification task.

Interestingly, the prompt variations for which we find significant differences largely correspond to those
which showed the largest distributional differences for LLM responses, with the sole exception of \textit{Emo-care},
which affected human responses much more than it did LLMs. Moreover, although human annotation results remain consistent across both tasks, certain prompt variants (e.g., \textit{Cap-label}, \textit{Def-both}, and \textit{Conf}) elicit markedly greater sensitivity in LLM performance on the offensiveness rating task than on the emotion classification task.

\subsection{RQ3: Which types of prompt variants yield similar responses between humans and LLMs?}

To explore the prompt variants that lead to similar response between humans and LLMs, we analyze the distributional distance between LLM-generated and human annotations, across prompt variants evaluated for the offensiveness rating and emotion classification tasks in our human study. Heatmaps presented in Figure~\ref{fig:off-matrix} (offensiveness rating) and Figure~\ref{fig:emo-matrix} (emotion classification) illustrate the pairwise distributional differences between responses from each $(P_i(\text{LLM}_k), P_j(\text{human}))$ pair, where $P_i$ and $P_j$ denote specific prompt variants and $\text{LLM}_k$ denotes one of the tested models.

For both tasks, we observe that LLaMA-3.1-8B and Falcon3-7B generally produces output distributions that are more aligned with human responses compared to other LLMs. 
This suggests that smaller LLMs may outperform larger models in approximating human behavior for some tasks.

Our analysis reveals that certain prompt modifications significantly increase the divergence in the distribution between LLM outputs and human annotations. In the offensiveness rating task, for example, the \textit{LabelOrd-rev} prompt variant leads to notable deviations in outputs generated by Mixtral-8x7B and Mistral-7B, while the \textit{AltLab-gran-2} variant causes pronounced discrepancies for the Llama models. Similarly, in the emotion classification task, the \textit{AltLab-keep-2} and \textit{Typo-label-3} variants tend to introduce higher divergence from human labels across most evaluated LLMs. Surprisingly, the \textit{Exp} variant demonstrates that requesting an explanation during annotation enhances the alignment between LLMs and humans.

Conversely, we observe stronger alignment between LLM and human output distributions when both are presented with identical prompts, as opposed to mismatched prompt variants. This effect is particularly evident for the \textit{AltLab-gran-2} and \textit{Def-both} variants in the offensiveness rating task, especially with LLaMA-3.1-8B, LLaMA-3.3-70B, and Mistral-7B. Additionally, in the emotion classification task, the \textit{Typo-label} and \textit{Exp} variants yield relatively consistent alignment across almost all evaluated LLMs, which exhibits less divergence.

\begin{table}[t]
\begin{tabular}{llS[table-format=5.2, parse-numbers=false]}
\toprule
Task & Prompt Variant & {$\chi^2$}\\
\midrule
\multirow{9}{2cm}{Offensiveness rating}
& LabelOrd-rev & 607.2*\\
& Syns-co-1 & (562.9)\\
& Typo-label-1 & (511.3) \\
& Cap-label & (529.4)\\
& AltLab-int-2 & 621.4** \\
& AltLab-gran-2 & 704.2***\\
& Def-both & 631.8**\\
& Conf & 612.2*\\
& Emo-care & (530.9)\\
\midrule
\multirow{9}{2cm}{Emotion classification}
& LabelOrd-shuff1 & (360.8)\\
& Syns-co-1 & (425.4)\\
& Typo-label-3 & 469.9**\\
& Cap-label & (404.8)\\
& AltLab-keep-2 & 766.7***\\
& Def-task & (372.9)\\
& Conf & (390.4)\\
& Exp & (427.2)\\
& Emo-care & 465.2*\\
\bottomrule
\end{tabular}
\caption{\label{tab:chi2} Aggregate $\chi^2$-test results for human responses.
In each case, we compare the human response distribution for a prompt variation against the response distribution
for the base prompt, in order to test which prompt variations affected human responses in a statistically significant manner.
Results reported are aggregated accross all instances, responses for which are assumed to be independent.
*: $p<0.05$, **: $p<0.01$, ***: $p<0.001$, (): $p \geq
0.05$}
\end{table}

\section{Conclusions and Future Work} \label{sec:conclude}

In this paper, we investigate whether humans exhibit prompt brittleness, a sensitivity to prompt variations, in a manner comparable to LLMs. Our findings demonstrate that, similar to LLMs, human responses can be influenced by specific prompt modifications, although humans display greater robustness to certain types of changes. To explore this, we develop a systematic method of prompt perturbations grounded in the hypothesis that prompt modifications fall into two categories: those that affect the distribution of human annotations and those that do not. 

Through human studies, we examine representative perturbations from each category. Notably, our analysis indicates that label substitutions induce comparable shifts in response distributions for both humans and LLMs. However, humans exhibit increased resilience to typographical errors and variations in label ordering where LLMs tend to struggle. 

Furthermore, our results suggest that when LLMs and humans are presented with identical prompts, their output distributions are more aligned than when using mismatched prompts. This indicates that using consistent prompt formulations across both groups can facilitate better alignment in annotation tasks.

Not all our findings are consistent across models. While this variability may reflect differences in LLM training or model sensitivity to the stochastic decoding strategy, we offer practical insights for prompt engineering and emphasize the value of prompt-aware annotation. 

Consequently, we advocate for future research to concentrate on objective tasks with lower inherent uncertainty than subjective tasks, to validate the generalizability of these results. Given the observed variability in LLM sensitivity to decoding strategies, further investigation into how decoding hyperparameters affect prompt brittleness is also warranted. Ultimately, our work
underscores the necessity of accounting for prompt brittleness when designing human annotation tasks and interpreting LLM outputs, encouraging greater attention to prompt formulation in annotation contexts.

\section*{Limitations}
While we explore all proposed categories of prompt variations, only a
subset of samples within each category can be evaluated through human annotation due to practical
constraints. Expanding the number
of tasks and instances would further improve the robustness of our
results. Among the various types of prompt perturbations investigated,
some variants remain underexplored. For instance, in the case of the
\textit{LabelOrd} modification, we do not test all possible
permutations of label orderings, which could limit the scope of our
conclusions.

Additionally, the computation of distance scores using Jensen–Shannon divergence is sensitive to the
size of the annotation distribution. In this study, the human annotations comprise only 10 samples, whereas the LLM-generated annotations comprise 100 samples. The relatively small human annotation set can result in higher variance in the distribution, which may increase the measured divergence compared to the larger LLM annotation set. Our comparisons between LLM-generated and human annotations could benefit from a more refined
calibration strategy to address the discrepancy in the number of
annotations per instance between the two sources.

Regarding the survey design, although we explicitly encourage
annotators to read the prompt for every instance, full control over
participant behavior is not feasible. It is possible that some
annotators may have ignored the prompt after becoming familiar with
the task structure.

Lastly, our experiments are conducted with only local
LLMs. Introducing commercial LLM APIs could enhance the granularity of
our study by contributing to real-world applications.  This would,
however, come at the cost of limited reproducability of the
experiments.

\section*{Ethical Considerations}

In our human study, all participants were informed that their
responses, including demographic information, would be used for a
scientific publication, and explicit consent was obtained. The
collected data is anonymized to protect participants'
privacy. Annotators who failed to pass one or more attention checks
were excluded from the final results. We follow the instructions and
suggestions of payment rate provided by the Prolific platform.

For the offensiveness rating task, participants were warned that the
content they would be exposed to might include offensive or explicit
language. In the emotion classification task involving prompts with
intentional typographical errors, participants were informed
afterwards that these typos were deliberately included for research
purposes, to avoid confusion or misinterpretation. We do not change
the intentional use for the datasets and models we refer to.

While one of the goals of this paper is to study the correlation of
variability between LLMs predictions and human annotations, we
acknowledge potential ethical concerns, particularly with regard to
bias in both LLM outputs and human judgments. Nevertheless, we believe
our findings provide valuable insights for future research in
automatic annotation processes using LLMs.

ChatGPT~\citep{chatgpt} was used as a tool to improve code generation
for figures and tables, as well as to assist with grammar and
vocabulary in the text of some sections of this paper.

\section*{Acknowledgments}

This work is funded by the project INPROMPT (Interactive Prompt
Optimization with the Human in the Loop for Natural Language
Understanding Model Development and Intervention, funded by the German
Research Foundation, KL 2869/13--1, project number 521755488).

\bibliography{custom}

\appendix

\section{Appendix}
\label{sec:appendix}

% \subsection{Heatmaps of Distance Scores between Prompt Variants} \label{app:heatmap}
% See Figure~\ref{fig:all-heatmaps}.

\subsection{Heatmaps of SpearmanR Statistics Across LLMs} \label{app:spearmanr}
\begin{figure*}[htbp]
  \centering
  
  \begin{subfigure}[t]{0.49\textwidth}
    \centering
    \includegraphics[width=\columnwidth]{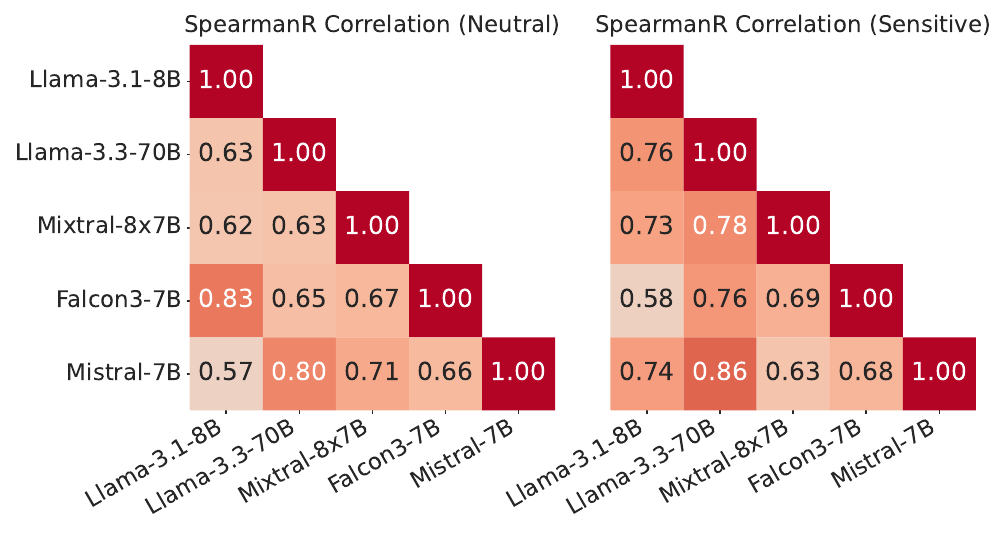}
    \caption{Offensiveness rating.}
    \label{fig:off-spearmanr}
  \end{subfigure}
  \hfill
  \begin{subfigure}[t]{0.49\textwidth}
    \centering
    \includegraphics[width=\columnwidth]{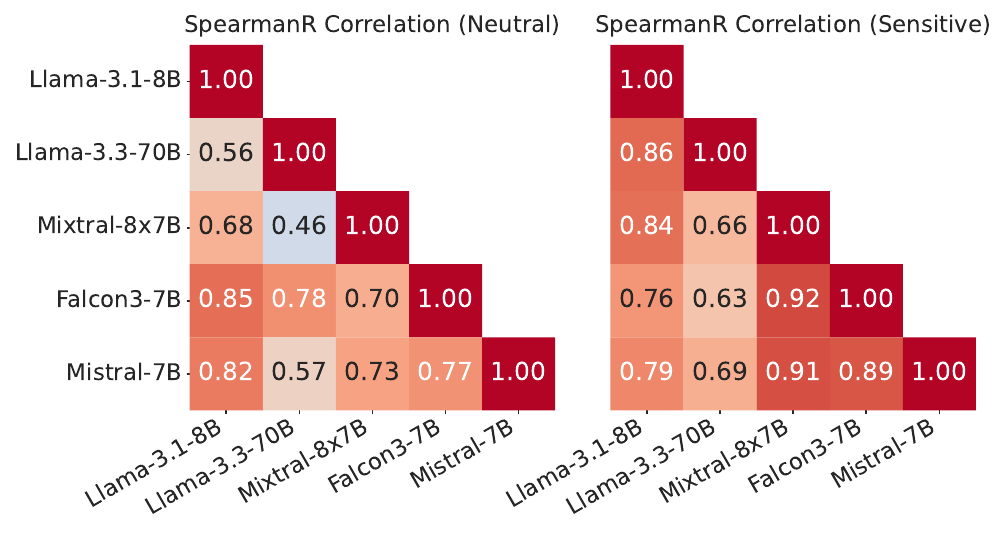}
    \caption{Politeness rating.}
    \label{fig:po-spearmanr}
  \end{subfigure}
  
  \vspace{0.3cm}
  
  \begin{subfigure}[t]{0.49\textwidth}
    \centering
    \includegraphics[width=\columnwidth]{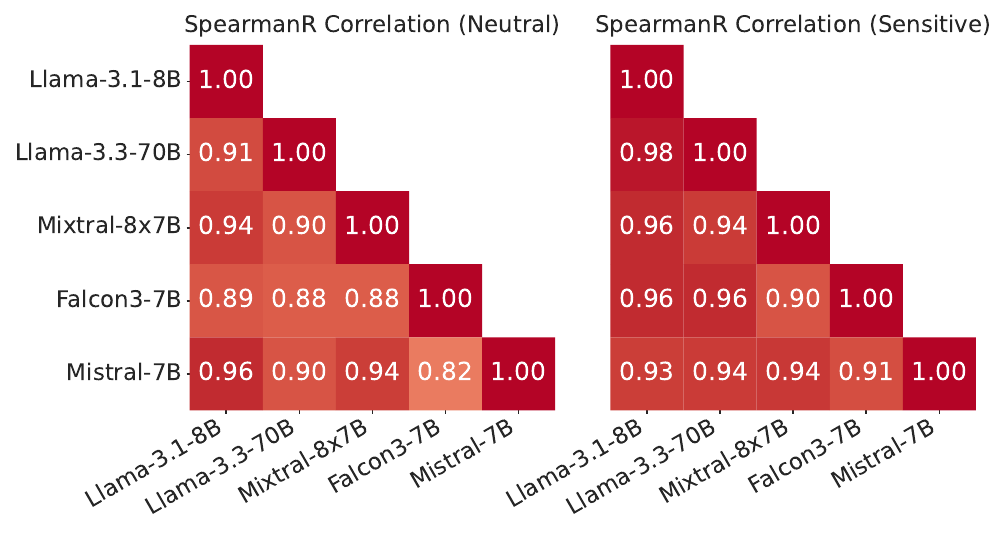}
    \caption{Emotion classification.}
    \label{fig:emo-spearmanr}
  \end{subfigure}
  \hfill
  \begin{subfigure}[t]{0.49\textwidth}
    \centering
    \includegraphics[width=\columnwidth]{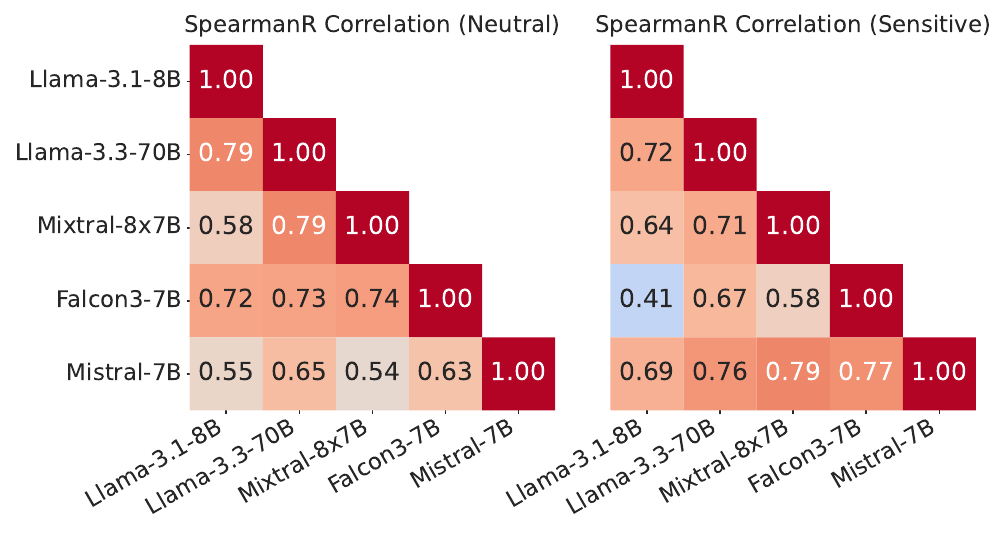}
    \caption{Irony detection.}
    \label{fig:iro-spearmanr}
  \end{subfigure}

  \caption{
    Heatmaps showing Spearman's rank correlation coefficients of distance scores for prompt perturbations across five LLMs. The four evaluated tasks are (a) offensiveness rating, (b) politeness rating, (c) emotion classification, and (d) irony detection. We present the results for prompt perturbations in the neutral and sensitive categories introduced in Section~\ref{sec:method}. Labels of x-axis and y-axis both refer to the names of LLMs. The coefficient values have a range from $-1$ to $1$, where $1$ means perfect monotonic increasing correlation (ranks agree exactly), -$1$ means monotonic decreasing correlation (ranks are opposites), $0$ means no monotonic relationship.
  }
  \label{fig:spearmanr}
\end{figure*}

We compute Spearman’s rank
correlation coefficients between pairs of models for the four evaluated tasks,
based on the ranking of Jensen--Shannon divergence scores induced by each prompt
variant. See Figure~\ref{fig:spearmanr}.

% \newpage
% \onecolumn
\subsection{Examples for Prompt Modifications} \label{app:prompts}

All prompt variations are derived from the `Base' prompt, based on different types of modifications introduced in Section~\ref{sec:method}. The placeholder \texttt{\{Inst.\}} denotes the instruction that constrains the LLM generation. In our experiments, \texttt{\{Inst.\}} is instantiated with the sentence: "Provide only the label and omit any justification." We provide all prompts used across the four tasks in our study: Table~\ref{tab:prompt-off} for offensiveness rating, Table~\ref{tab:prompt-po} for politeness rating, Table~\ref{tab:prompt-iro} for irony detection, and Table~\ref{tab:prompt-emo} for emotion classification, along with their corresponding modification types.

\subsection{Instructions of Survey for Human Study}
\begin{figure*}[htbp]
\centering
  \includegraphics[width=0.8\linewidth]{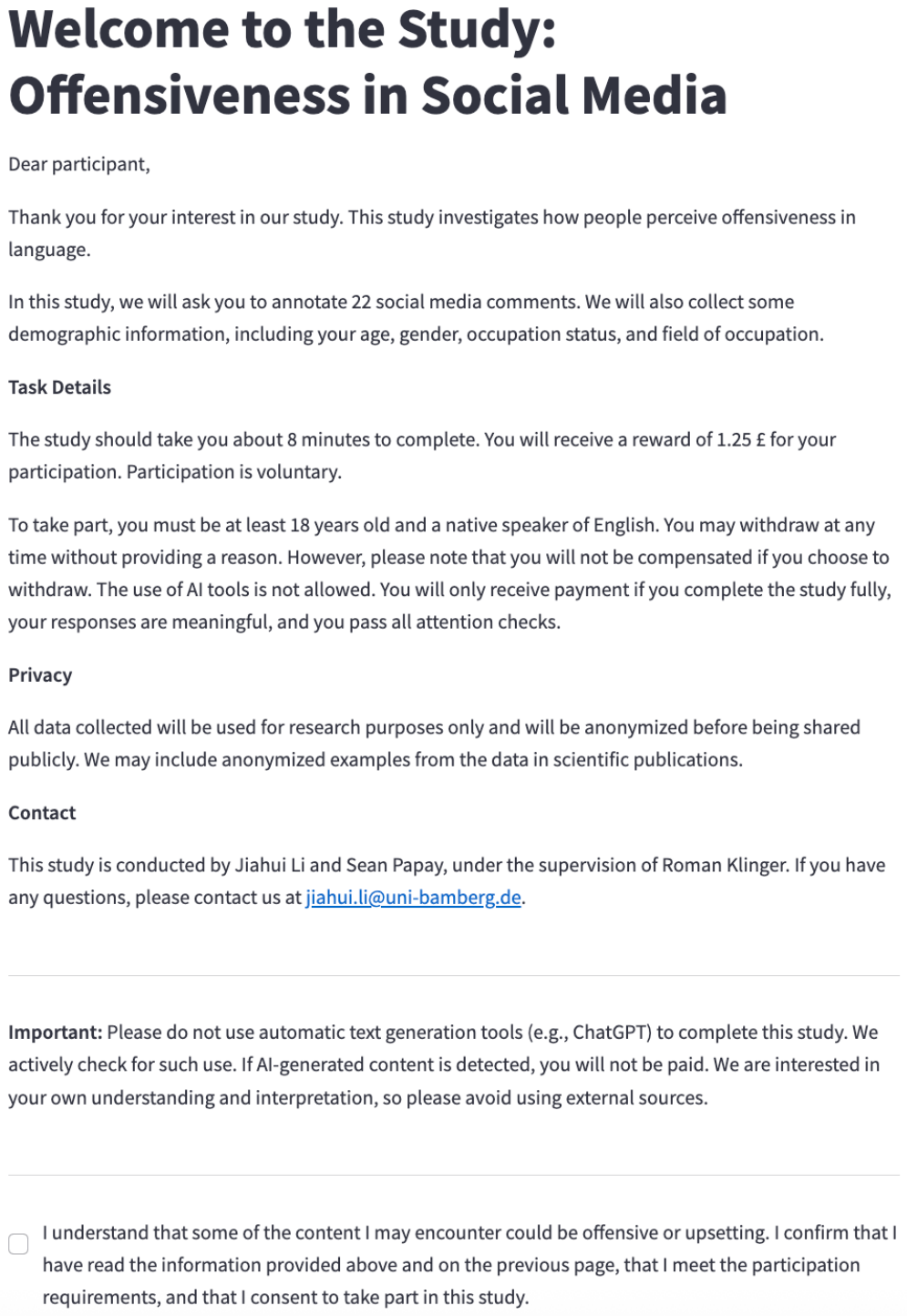} 
  % \hfill
  % \includegraphics[width=0.48\linewidth]{example-image-b}
  \caption {A screenshot of the instruction page for offensiveness rating survey.}
  \label{fig:screen-off}
\end{figure*}

\begin{figure*}[htbp]
\centering
  \includegraphics[width=0.8\linewidth]{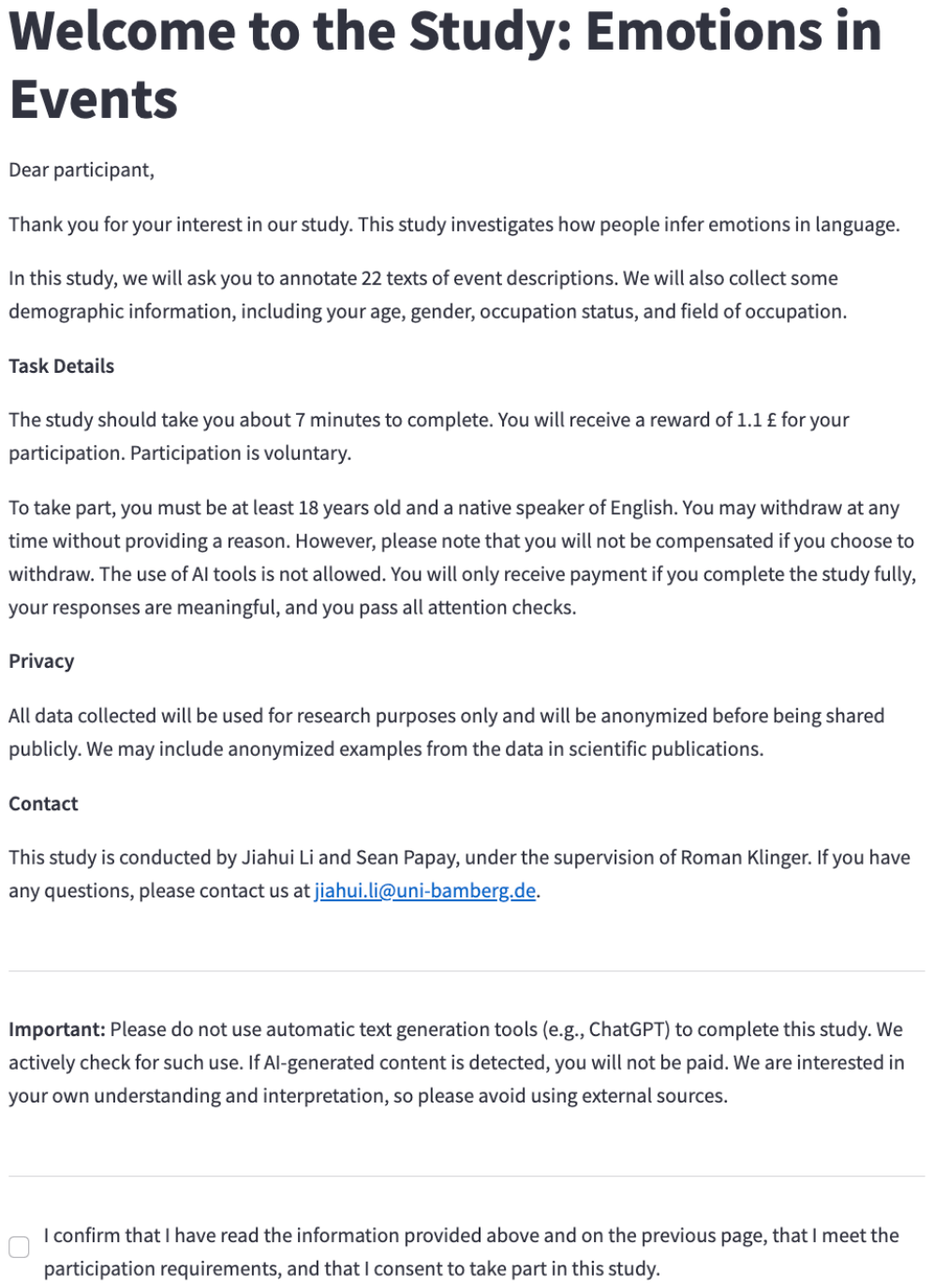} 
  \caption {A screenshot of the instruction page for the emotion classification survey.}
  \label{fig:screen-emo}
\end{figure*}

\begin{figure*}[htbp]
\centering
  \includegraphics[width=0.8\linewidth]{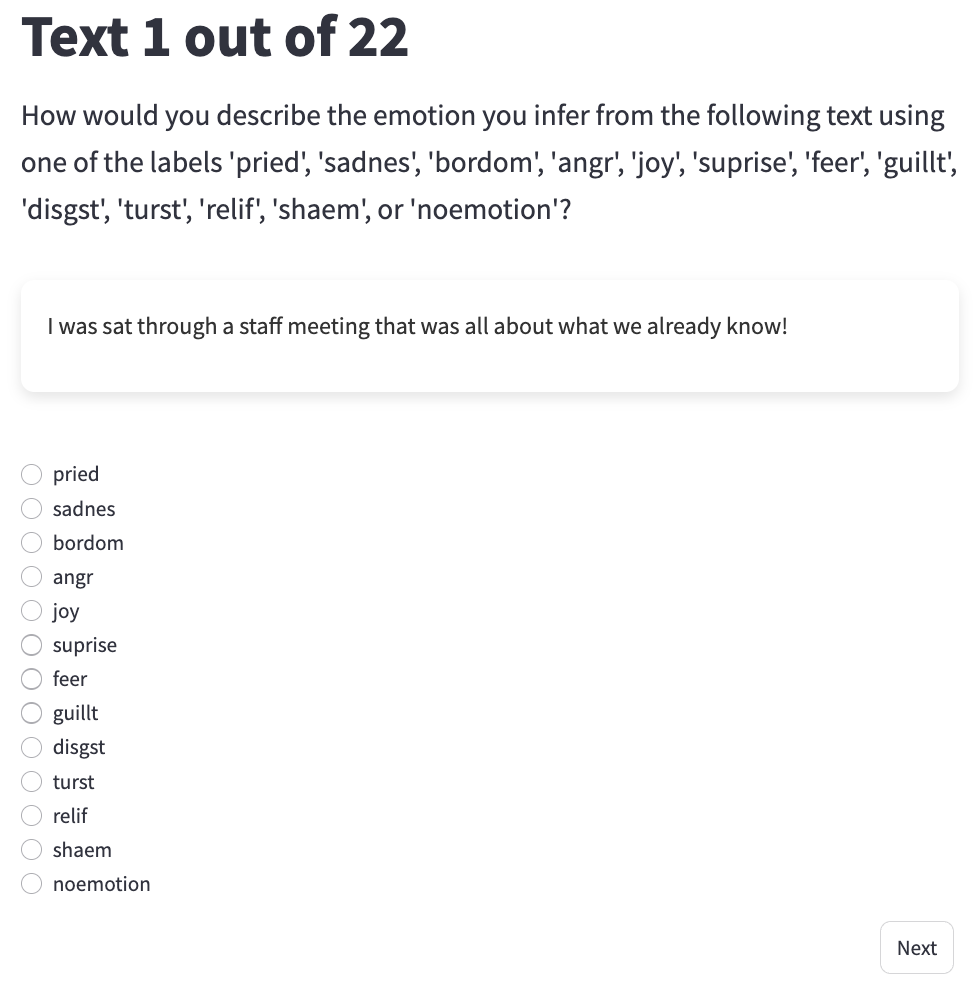} 
  \caption {A screenshot example of the annotation task for the emotion classification survey. The prompt variant is instantiated with a typological error modification (Typo-label-3).}
  \label{fig:screen-emotypo}
\end{figure*}

We provide the screenshots of the instruction page guiding our human study surveys. See Figure~\ref{fig:screen-off} for the offensiveness rating task and Figure~\ref{fig:screen-emo} for the emotion classification task. An example of the annotation case for the emotion classification task, which involves a prompt with a typological error modification, is shown in Figure~\ref{fig:screen-emotypo}.

\onecolumn
\begin{center}
\small
\begin{longtable}{p{2cm}p{12.5cm}}

\caption{Examples of prompt variants with different modification types used for the offensiveness rating task.}
\label{tab:prompt-off} \\

\toprule
\textbf{Type} & \textbf{Prompt} \\
\midrule
\endfirsthead

\multicolumn{2}{c}{{\tablename\ \thetable{} -- continued from previous page}} \\
\toprule
\textbf{Type} & \textbf{Prompt} \\
\midrule
\endhead

\midrule
\multicolumn{2}{r}{{Continued on next page}} \\
\bottomrule
\endfoot

\bottomrule
\endlastfoot

Base & How would you rate the offensiveness of the following text using one of the labels `not offensive at all', `slightly offensive', `moderately offensive', `offensive', or `very offensive'? \texttt{\{Inst.\}} \\
Imper & Rate the offensiveness of the following text using one of the labels `not offensive at all', `slightly offensive', `moderately offensive', `offensive', or `very offensive'. \texttt{\{Inst.\}} \\
Imper-pls & Please rate the offensiveness of the following text using one of the labels `not offensive at all', `slightly offensive', `moderately offensive', `offensive', or `very offensive'. \texttt{\{Inst.\}} \\
LabelOrd-rev & How would you rate the offensiveness of the following text using one of the labels `very offensive', `offensive', `moderately offensive', `slightly offensive', or `not offensive at all'? \texttt{\{Inst.\}} \\
LabelPos-start & Using one of the labels `not offensive at all', `slightly offensive', `moderately offensive', `offensive', or `very offensive', how would you rate the offensiveness of the following text? \texttt{\{Inst.\}} \\
Syns-verb-1 & How do you rate the offensiveness of the following text using one of the labels `not offensive at all', `slightly offensive', `moderately offensive', `offensive', or `very offensive'? \texttt{\{Inst.\}} \\
Syns-verb-2 & How would you evaluate the offensiveness of the following text using one of the labels `not offensive at all', `slightly offensive', `moderately offensive', `offensive', or `very offensive'? \texttt{\{Inst.\}} \\
Syns-verb-3 & How would you judge the offensiveness of the following text using one of the labels `not offensive at all', `slightly offensive', `moderately offensive', `offensive', or `very offensive'? \texttt{\{Inst.\}} \\
Syns-noun & How would you rate the offensiveness of the following text on the scale `not offensive at all', `slightly offensive', `moderately offensive', `offensive', or `very offensive'? \texttt{\{Inst.\}} \\
Syns-prep-1 & How would you rate the offensiveness of the following text according to the labels `not offensive at all', `slightly offensive', `moderately offensive', `offensive', or `very offensive'? \texttt{\{Inst.\}} \\
Syns-prep-2 & How would you rate the offensiveness of the following text given the labels `not offensive at all', `slightly offensive', `moderately offensive', `offensive', or `very offensive'? \texttt{\{Inst.\}} \\
Syns-co-1 & How would you rate the offensiveness of the following text using one of the labels `not offensive at all', `slightly offensive', `moderately offensive', `offensive', and `very offensive'? \texttt{\{Inst.\}} \\
Syns-co-2 & How would you rate the offensiveness of the following text using one of the labels `not offensive at all', `slightly offensive', `moderately offensive', `offensive', `very offensive'? \texttt{\{Inst.\}} \\
Typo-task-1 & How would you raet the offensiveness of the following text using one of the labels `not offensive at all', `slightly offensive', `moderately offensive', `offensive',`very offensive'? \texttt{\{Inst.\}} \\
Typo-task-2 & How would you rate the offensivness of the following text using one of the labels `not offensive at all', `slightly offensive', `moderately offensive', `offensive', or `very offensive'? \texttt{\{Inst.\}} \\
Typo-task-3 & How would you rate the offensiveness of the following text using one of the lables `not offensive at all', `slightly offensive', `moderately offensive', `offensive', or `very offensive'? \texttt{\{Inst.\}} \\
Typo-label-1 & How would you rate the offensiveness of the following text using one of the labels `not offensiv at all', `slightly offensiv', `moderately offensiv', `offensiv', or `very offensiv'? \texttt{\{Inst.\}} \\
Typo-label-2 & How would you rate the offensiveness of the following text using one of the labels `not offensive at al', `slightly offensive', `moderatly offensive', `offensive', or `very offensive'? \texttt{\{Inst.\}} \\
Cap-task & How would you rate the OFFENSIVENESS of the following TEXT using one of the labels `not offensive at all', `slightly offensive', `moderately offensive', `offensive', or `very offensive'? \texttt{\{Inst.\}} \\
Cap-label & How would you rate the offensiveness of the following text using one of the labels `NOT OFFENSIVE AT ALL', `SLIGHTLY OFFENSIVE', `MODERATELY OFFENSIVE', `OFFENSIVE', or `VERY OFFENSIVE'? \texttt{\{Inst.\}} \\
PM-1 & How would you rate the offensiveness of the following text using one of the labels not offensive at all, slightly offensive, moderately offensive, offensive, or very offensive? \texttt{\{Inst.\}} \\
PM-2 & How would you rate the offensiveness of the following text using one of the labels: `not offensive at all', `slightly offensive', `moderately offensive', `offensive', or `very offensive'? \texttt{\{Inst.\}} \\
PM-3 & How would you rate the offensiveness of the following text using one of the labels `not offensive at all'; `slightly offensive'; `moderately offensive'; `offensive'; or `very offensive'? \texttt{\{Inst.\}} \\
AltLab-keep-1 & How would you rate the offensiveness of the following text using one of the labels `not offensive at all', `mildly offensive', `fairly offensive', `quite offensive', or `very offensive'? \texttt{\{Inst.\}} \\
AltLab-keep-2 & How would you rate the offensiveness of the following text on a Likert scale from 1-5, where 1 means `not offensive at all' and 5 means `very offensive'? \texttt{\{Inst.\}} \\
AltLab-keep-3 & How would you rate the offensiveness of the following text on a Likert scale from 1-5, where 1 means `not offensive at all', 2 means `slightly offensive', 3 means `moderately offensive', 4 means `offensive', and 5 means `very offensive'? \texttt{\{Inst.\}} \\
AltLab-keep-4 & How would you rate the offensiveness of the following text on a Likert scale from 0-4, where 0 means 'not offensive at all' and 4 means 'very offensive'? \texttt{\{Inst.\}} \\
AltLab-int-1 & How would you rate the offensiveness of the following text using one of the labels `not offensive at all', `slightly offensive', `moderately offensive', `very offensive', or `extremely offensive'? \texttt{\{Inst.\}} \\
AltLab-int-2 & How would you rate the offensiveness of the following text using one of the labels `not offensive', `mild', `moderate', `strong', or `extreme'? \texttt{\{Inst.\}} \\
AltLab-int-3 & How would you rate the offensiveness of the following text using one of the labels `no offense', `low', `medium', `high', or `severe'? \texttt{\{Inst.\}} \\
AltLab-gran-1 & How would you rate the offensiveness of the following text on a Likert scale from 1-4, where 1 means `not offensive at all' and 4 means `very offensive'? \texttt{\{Inst.\}} \\
AltLab-gran-2 & How would you rate the offensiveness of the following text on a Likert scale from 1-7, where 1 means `not offensive at all' and 7 means `very offensive'? \texttt{\{Inst.\}} \\
AltLab-gran-3 & How would you rate the offensiveness of the following text on a Likert scale from 1-10, where 1 means `not offensive at all' and 10 means `very offensive'? \texttt{\{Inst.\}} \\

Def-task & How would you rate the offensiveness of the following text using one of the labels `not offensive at all', `slightly offensive', `moderately offensive', `offensive', or `very offensive'? Offensiveness refers to how likely the text is to be perceived as inappropriate, insulting, or harmful in a general social context. Consider factors such as language, tone, and potential harm to individuals or groups. \texttt{\{Inst.\}} \\
Def-label-1 & How would you rate the offensiveness of the following text using one of these labels?\newline`not offensive at all': The text contains no language or content that is likely to be considered offensive or inappropriate in most social and cultural contexts.\newline`slightly offensive': The text may contain mildly inappropriate language or implications, but it is unlikely to cause strong negative reactions.\newline`moderately offensive': The text includes content that some people might find offensive or inappropriate, but it is not explicitly harmful or strongly insulting.\newline`offensive': The text contains language or ideas that many people would consider inappropriate, offensive, or disrespectful.\newline`very offensive': The text includes highly inappropriate, harmful, or offensive content that is likely to cause strong negative reactions or be considered unacceptable in most contexts.\newline\texttt{\{Inst.\}} \\
Def-label-2 & How would you rate the offensiveness of the following text using one of these labels?\newline`not offensive at all': The text contains no language or content that could reasonably upset, insult, or demean any individual or group. It is neutral or positive in tone.\newline`slightly offensive': The text includes mildly insensitive or potentially provocative language, but the intent is ambiguous or the impact is minimal.\newline`moderately offensive': The text contains clearly disrespectful or derogatory content, though not extreme. It may offend some audiences but not others.\newline`offensive': The text uses overtly harmful language, slurs, or stereotypes likely to upset most readers. Intent may be hostile or careless.\newline`very offensive': The text includes severe slurs, threats, dehumanizing language, or hate speech that would widely be considered unacceptable.\newline\texttt{\{Inst.\}} \\
Def-both & How would you rate the offensiveness of the following text using one of these labels? Offensiveness refers to how likely the text is to be perceived as inappropriate, insulting, or harmful in a general social context. Consider factors such as language, tone, and potential harm to individuals or groups.\newline`not offensive at all': The text contains no language or content that is likely to be considered offensive or inappropriate in most social and cultural contexts.\newline`slightly offensive': The text may contain mildly inappropriate language or implications, but it is unlikely to cause strong negative reactions.\newline`moderately offensive': The text includes content that some people might find offensive or inappropriate, but it is not explicitly harmful or strongly insulting.\newline`offensive': The text contains language or ideas that many people would consider inappropriate, offensive, or disrespectful.\newline`very offensive': The text includes highly inappropriate, harmful, or offensive content that is likely to cause strong negative reactions or be considered unacceptable in most contexts.\newline\texttt{\{Inst.\}} \\
Conf & How would you rate the offensiveness of the following text using one of the labels `not offensive at all', `slightly offensive', `moderately offensive', `offensive', or `very offensive'? Provide the label along with a confidence score (1-5) of your rating, and omit any justification. \\
Exp & How would you rate the offensiveness of the following text using one of the labels `not offensive at all', `slightly offensive', `moderately offensive', `offensive', or `very offensive'? Provide the label along with a justification for your rating. \\
Emo-1 & How would you rate the offensiveness of the following text using one of the labels `not offensive at all', `slightly offensive', `moderately offensive', `offensive', or `very offensive'? Important: Trust your gut reaction! \texttt{\{Inst.\}} \\
Emo-2 & How would you rate the offensiveness of the following text using one of the labels `not offensive at all', `slightly offensive', `moderately offensive', `offensive', or `very offensive'? Warning: Some texts may be disturbing. \texttt{\{Inst.\}} \\
Emo-3 & How would you rate the offensiveness of the following text using one of the labels `not offensive at all', `slightly offensive', `moderately offensive', `offensive', or `very offensive'? Please be careful with your rating. \texttt{\{Inst.\}} \\
Emo-4 & How would you rate the offensiveness of the following text using one of the labels `not offensive at all', `slightly offensive', `moderately offensive', `offensive', or `very offensive'? Important: Choose quickly and go with your first reaction! \texttt{\{Inst.\}} \\

\end{longtable}
\end{center}
\twocolumn

\onecolumn
\begin{center}
\small
\begin{longtable}{p{2cm}p{12.5cm}}
\caption{Examples of prompt variants with different modification types used for the politeness rating task.}
\label{tab:prompt-po} \\

\toprule
\textbf{Type} & \textbf{Prompt} \\
\midrule
\endfirsthead

\multicolumn{2}{c}{{\tablename\ \thetable{} -- continued from previous page}} \\
\toprule
\textbf{Type} & \textbf{Prompt} \\
\midrule
\endhead

\midrule
\multicolumn{2}{r}{{Continued on next page}} \\
\bottomrule
\endfoot

\bottomrule
\endlastfoot

Base & How would you rate the politeness of the following text using one of the labels `not polite at all', `slightly polite', `moderately polite', `polite', or `very polite'? \texttt{\{Inst.\}} \\
Imper & Rate the politeness of the following text using one of the labels `not polite at all', `slightly polite', `moderately polite', `polite', or `very polite'. \texttt{\{Inst.\}} \\
Imper-pls & Please rate the politeness of the following text using one of the labels `not polite at all', `slightly polite', `moderately polite', `polite', or `very polite'. \texttt{\{Inst.\}} \\
LabelOrd-rev & How would you rate the politeness of the following text using one of the labels `very polite', `polite', `moderately polite', `slightly polite', or `not polite at all'? \texttt{\{Inst.\}} \\
LabelPos-start & Using one of the labels `not polite at all', `slightly polite', `moderately polite', `polite', or `very polite', how would you rate the politeness of the following text? \texttt{\{Inst.\}} \\
Syns-verb-1 & How do you rate the politeness of the following text using one of the labels `not polite at all', `slightly polite', `moderately polite', `polite', or `very polite'? \texttt{\{Inst.\}} \\
Syns-verb-2 & How would you evaluate the politeness of the following text using one of the labels `not polite at all', `slightly polite', `moderately polite', `polite', or `very polite'? \texttt{\{Inst.\}} \\
Syns-verb-3 & How would you judge the politeness of the following text using one of the labels `not polite at all', `slightly polite', `moderately polite', `polite', or `very polite'? \texttt{\{Inst.\}} \\
Syns-noun & How would you rate the politeness of the following text on the scale `not polite at all', `slightly polite', `moderately polite', `polite', or `very polite'? \texttt{\{Inst.\}} \\
Syns-prep-1 & How would you rate the politeness of the following text according to the labels `not polite at all', `slightly polite', `moderately polite', `polite', or `very polite'? \texttt{\{Inst.\}} \\
Syns-prep-2 & How would you rate the politeness of the following text given the labels `not polite at all', `slightly polite', `moderately polite', `polite', or `very polite'? \texttt{\{Inst.\}} \\
Syns-co-1 & How would you rate the politeness of the following text using one of the labels `not polite at all', `slightly polite', `moderately polite', `polite', and `very polite'? \texttt{\{Inst.\}} \\
Syns-co-2 & How would you rate the politeness of the following text using one of the labels `not polite at all', `slightly polite', `moderately polite', `polite', `very polite'? \texttt{\{Inst.\}} \\
Typo-task-1 & How would you raet the politeness of the following text using one of the labels `not polite at all', `slightly polite', `moderately polite', `polite',`very polite'? \texttt{\{Inst.\}} \\
Typo-task-2 & How would you rate the politness of the following text using one of the labels `not polite at all', `slightly polite', `moderately polite', `polite', or `very polite'? \texttt{\{Inst.\}} \\
Typo-task-3 & How would you rate the politeness of the following text using one of the lables `not polite at all', `slightly polite', `moderately polite', `polite', or `very polite' \texttt{\{Inst.\}} \\
Typo-label-1 & How would you rate the politeness of the following text using one of the labels `not polit at all', `slightly polit', `moderately polit', `polit', or `very polit'? \texttt{\{Inst.\}} \\
Typo-label-2 & How would you rate the politeness of the following text using one of the labels `not polite at al', `slightly polite', `moderatly polite', `polite', or `very polite' \texttt{\{Inst.\}} \\
Cap-task & How would you rate the POLITENESS of the following TEXT using one of the labels `not polite at all', `slightly polite', `moderately polite', `polite', or `very polite'? \texttt{\{Inst.\}} \\
Cap-label & How would you rate the politeness of the following text using one of the labels `NOT POLITE AT ALL', `SLIGHTLY POLITE', `MODERATELY POLITE', `POLITE', or `VERY POLITE' \texttt{\{Inst.\}} \\
PM-1 & How would you rate the politeness of the following text using one of the labels not polite at all, slightly polite, moderately polite, polite, or very polite? \texttt{\{Inst.\}} \\
PM-2 & How would you rate the politeness of the following text using one of the labels: `not polite at all', `slightly polite', `moderately polite', `polite', or `very polite'? \texttt{\{Inst.\}} \\
PM-3 & How would you rate the politeness of the following text using one of the labels `not polite at all'; `slightly polite'; `moderately polite'; `polite'; or `very polite'? \texttt{\{Inst.\}} \\
AltLab-keep-1 & How would you rate the politeness of the following text using one of the labels `not polite at all', `mildly polite', `fairly polite', `quite polite', or `very polite'? \texttt{\{Inst.\}} \\
AltLab-keep-2 & How would you rate the politeness of the following text on a Likert scale from 1-5, where 1 means `not polite at all' and 5 means `very polite'? \texttt{\{Inst.\}} \\
AltLab-keep-3 & How would you rate the politeness of the following text on a Likert scale from 1-5, where 1 means `not polite at all', 2 means `slightly polite', 3 means `moderately polite', 4 means `polite', and 5 means `very polite'? \texttt{\{Inst.\}} \\
AltLab-keep-4 & How would you rate the politeness of the following text on a Likert scale from 0-4, where 0 means `not polite at all' and 4 means `very polite'? \texttt{\{Inst.\}} \\
AltLab-int-1 & How would you rate the politeness of the following text using one of the labels `not polite at all', `slightly polite', `moderately polite', `very polite', or 'extremely polite'? \texttt{\{Inst.\}} \\
AltLab-int-2 & How would you rate the politeness of the following text using one of the labels `not polite', `mild', `moderate', `strong', or `extreme'? \texttt{\{Inst.\}} \\
AltLab-int-3 & How would you rate the politeness of the following text using one of the labels `very impolite', `impolite', `neutral', `polite', `very polite' \texttt{\{Inst.\}} \\
AltLab-gran-1 & How would you rate the politeness of the following text on a Likert scale from 1-4, where 1 means `not polite at all' and 4 means `very polite'? \texttt{\{Inst.\}} \\
AltLab-gran-2 & How would you rate the politeness of the following text on a Likert scale from 1-7, where 1 means `not polite at all' and 7 means `very polite'? \texttt{\{Inst.\}} \\
AltLab-gran-3 & How would you rate the politeness of the following text on a Likert scale from 1-10, where 1 means `not polite at all' and 10 means `very polite'? \texttt{\{Inst.\}} \\
Def-task & How would you rate the politeness of the following text using one of the labels `not polite at all', `slightly polite', `moderately polite', `polite', or `very polite'? Assessment is based on its choice of words, tone, and overall manner of expression. Consider whether the text demonstrates courtesy, respect, and appropriateness in communication. \texttt{\{Inst.\}} \\
Def-label-1 & How would you rate the politeness of the following text using one of these labels?
\newline`not polite at all': The text is rude, offensive, lacks any form of courtesy. It may include insults, aggressive language, a dismissive tone.
\newline`slightly polite': The text is somewhat courteous but may contain blunt phrasing, minor rudeness, a lack of warmth. It is not outright offensive but could be perceived as unfriendly.
\newline`moderately polite': The text maintains a neutral to mildly respectful tone. It avoids harsh language but does not go out of its way to be particularly warm courteous.
\newline`polite': The text is respectful, considerate, and maintains a generally pleasant tone. It avoids any harsh dismissive language and follows social norms of politeness.
\newline`very polite': The text is exceptionally courteous, warm, and respectful. It may include formal expressions of gratitude, apologies, other markers of strong politeness.\newline\texttt{\{Inst.\}} \\
Def-label-2 & How would you rate the politeness of the following text using one of these labels?
\newline`not polite at all': The text lacks any politeness; it may be rude, blunt, or offensive.
\newline`slightly polite': Minimal politeness.
\newline`moderately polite': Clearly polite but neutral or formal; not overly warm.
\newline`polite': Consistently courteous, with clear respect and positive tone.
\newline`very polite': Exceptionally warm, respectful, or deferential; may include extra softening phrases.\newline\texttt{\{Inst.\}} \\
Def-both & How would you rate the politeness of the following text using one of the labels `not polite at all', `slightly polite', `moderately polite', `polite', or `very polite'? Assessment is based on its choice of words, tone, and overall manner of expression. Consider whether the text demonstrates courtesy, respect, and appropriateness in communication.
\newline`not polite at all': The text lacks any politeness; it may be rude, blunt, or offensive.
\newline`slightly polite': Minimal politeness.
\newline`moderately polite': Clearly polite but neutral or formal; not overly warm.
\newline`polite': Consistently courteous, with clear respect and positive tone.
\newline`very polite': Exceptionally warm, respectful, or deferential; may include extra softening phrases.\newline\texttt{\{Inst.\}} \\
Conf & How would you rate the politeness of the following text using one of the labels `not polite at all', `slightly polite', `moderately polite', `polite', or `very polite'? Provide the label along with a confidence score (1-5) of your rating, and omit any justification. \\
Exp & How would you rate the politeness of the following text using one of the labels `not polite at all', `slightly polite', `moderately polite', `polite', or `very polite'? Provide the label along with a justification for your rating. \\
Emo-1 & How would you rate the politeness of the following text using one of the labels `not polite at all', `slightly polite', `moderately polite', `polite', or `very polite'? Important: Trust your gut reaction! \texttt{\{Inst.\}} \\
Emo-2 & How would you rate the politeness of the following text using one of the labels `not polite at all', `slightly polite', `moderately polite', `polite', or `very polite'? Warning: Some texts may be disturbing. \texttt{\{Inst.\}} \\
Emo-3 & How would you rate the politeness of the following text using one of the labels `not polite at all', `slightly polite', `moderately polite', `polite', or `very polite'? Please be careful with your rating. \texttt{\{Inst.\}} \\
Emo-4 & How would you rate the politeness of the following text using one of the labels `not polite at all', `slightly polite', `moderately polite', `polite', or `very polite'? Important: Choose quickly and go with your first reaction! \texttt{\{Inst.\}} \\

\end{longtable}
\end{center}
\twocolumn

\onecolumn
\begin{center}
\small
\begin{longtable}{p{2cm}p{12.5cm}}
\caption{Examples of prompt variants with different modification types used for the irony detection task.}
\label{tab:prompt-iro} \\

\toprule
\textbf{Type} & \textbf{Prompt} \\
\midrule
\endfirsthead

\multicolumn{2}{c}{{\tablename\ \thetable{} -- continued from previous page}} \\
\toprule
\textbf{Type} & \textbf{Prompt} \\
\midrule
\endhead

\midrule
\multicolumn{2}{r}{{Continued on next page}} \\
\bottomrule
\endfoot

\bottomrule
\endlastfoot

Base & How would you classify the following reply to the given message using one of the labels `ironic' or `not ironic'? \texttt{\{Inst.\}} \\
Imper & Classify the following reply to the given message using one of the labels `ironic' or `not ironic'. \texttt{\{Inst.\}} \\
Imper-pls & Please classify the following reply to the given message using one of the labels `ironic' or `not ironic'. \texttt{\{Inst.\}} \\
LabelOrd-rev & How would you classify the following reply to the given message using one of the labels `not ironic' or `ironic'? \texttt{\{Inst.\}} \\
LabelPos-start & Using one of the labels `ironic' or `not ironic', how would you classify the following reply to the given message? \texttt{\{Inst.\}} \\
Syns-verb-1 & How would you assess the following reply to the given message using one of the labels `ironic' or `not ironic'? \texttt{\{Inst.\}} \\
Syns-verb-2 & How would you categorize the following reply to the given message using one of the labels `ironic' or `not ironic'? \texttt{\{Inst.\}} \\
Syns-verb-3 & How would you identify the following reply to the given message using one of the labels `ironic' or `not ironic'? \texttt{\{Inst.\}} \\
Syns-noun & How would you classify the following reply to the given message using one of the categories `ironic' or `not ironic'? \texttt{\{Inst.\}} \\
Syns-prep-1 & How would you classify the following reply to the given message according to the labels `ironic' or `not ironic'? \texttt{\{Inst.\}} \\
Syns-prep-2 & How would you classify the following reply to the given message given the labels `ironic' or `not ironic'? \texttt{\{Inst.\}} \\
Syns-co-1 & How would you classify the following reply to the given message using one of the labels `ironic' and `not ironic'? \texttt{\{Inst.\}} \\
Syns-co-2 & How would you classify the following reply to the given message using one of the labels `ironic', `not ironic'? \texttt{\{Inst.\}} \\
Typo-task-1 & How would you clasify the following reply to the given message using one of the labels `ironic' or `not ironic'? \texttt{\{Inst.\}} \\
Typo-task-2 & How would you classify the following reply to the given mesage using one of the labels `ironic' or `not ironic'? \texttt{\{Inst.\}} \\
Typo-task-3 & How would you classify the following reply to the given message using one of the lables `ironic' or `not ironic'? \texttt{\{Inst.\}} \\
Typo-label-1 & How would you classify the following reply to the given message using one of the labels `irnoic' or `not irnoic'? \texttt{\{Inst.\}} \\

Typo-label-2 & How would you classify the following reply to the given message using one of the labels `ironc' or `not ironc'? \texttt{\{Inst.\}} \\
Typo-label-3 & How would you classify the following reply to the given message using one of the labels `ironik' or `not ironik'? \texttt{\{Inst.\}} \\
Cap-task & How would you CLASSIFY the following REPLY to the given MESSAGE using one of the labels `ironic' or `not ironic'? \texttt{\{Inst.\}} \\
Cap-label & How would you classify the following reply to the given message using one of the labels `IRONIC' or `NOT IRONIC'? \texttt{\{Inst.\}} \\
PM-1 & How would you classify the following reply to the given message using one of the labels ironic or not ironic? \texttt{\{Inst.\}} \\
PM-2 & How would you classify the following reply to the given message using one of the labels: `ironic' or `not ironic'? \texttt{\{Inst.\}} \\
PM-3 & How would you classify the following reply to the given message using one of the labels `ironic'; `not ironic'? \texttt{\{Inst.\}} \\
AltLab-keep-1 & How would you classify the following reply to the given message using one of the labels `ironic' or `literal'? \texttt{\{Inst.\}} \\
AltLab-keep-2 & How would you classify the following reply to the given message using one of the labels `ironic' or `neutral'? \texttt{\{Inst.\}} \\
AltLab-keep-3 & How would you classify the following reply to the given message using one of the binary labels 0 (ironic) or 1 (not ironic)? \texttt{\{Inst.\}} \\
AltLab-keep-4 & How would you classify the following reply to the given message using one of the binary labels 1 (ironic) or 0 (not ironic)? \texttt{\{Inst.\}} \\
AltLab-int-1 & How would you classify the following reply to the given message using one of the labels `very ironic' or `not ironic at all'? \texttt{\{Inst.\}} \\
AltLab-int-2 & How would you classify the following reply to the given message using one of the labels `negative' (ironic) or `positive' (not ironic)? \texttt{\{Inst.\}} \\
AltLab-gran & How would you classify the following reply to the given message using one of the labels `explicitly ironic', `implicitly ironic', or `not ironic'? \texttt{\{Inst.\}} \\
Def-task & How would you classify the following reply to the given message using one of the labels `ironic' or `not ironic'? Irony implies a meaning that contrasts with its literal wording, often to express sarcasm, criticism, or humor. \texttt{\{Inst.\}} \\
Def-label-1 & How would you classify the following reply to the given message using one of these labels?
        \newline`ironic': The reply expresses a meaning that is intentionally contrary to its surface content.
        \newline`not ironic': The reply is direct, sincere, and does not involve irony.
        \newline\texttt{\{Inst.\}} \\
Def-label-2 & How would you classify the following reply to the given message using one of these labels?
        \newline`ironic': The reply says one thing but means something else (usually the opposite).
        \newline`not ironic': The reply is straightforward and honest without hidden meaning—the words match the intent.
        \newline\texttt{\{Inst.\}} \\
Def-both & How would you classify the following reply to the given message using one of these labels? Irony implies a meaning that contrasts with its literal wording, often to express sarcasm, criticism, or humor.
        \newline`ironic': The reply says one thing but means something else (usually the opposite).
        \newline`not ironic': The reply is straightforward and honest without hidden meaning—the words match the intent.
        \newline\texttt{\{Inst.\}} \\
Conf & How would you classify the following reply to the given message using one of the labels `ironic' or `not ironic'? Provide the label along with a confidence score (1-5) of your classification, and omit any justification. \\
Exp & How would you classify the following reply to the given message using one of the labels `ironic' or `not ironic'? Provide the label along with a justification for your classification. \\
Emo-1 & How would you classify the following reply to the given message using one of the labels `ironic' or `not ironic'? Important: Trust your gut reaction! \texttt{\{Inst.\}} \\
Emo-2 & How would you classify the following reply to the given message using one of the labels `ironic' or `not ironic'? Warning: Some texts may be disturbing. \texttt{\{Inst.\}} \\
Emo-3 & How would you classify the following reply to the given message using one of the labels `ironic' or `not ironic'?Please be careful with your classification. \texttt{\{Inst.\}} \\
Emo-4 & How would you classify the following reply to the given message using one of the labels `ironic' or `not ironic'? Important: Choose quickly and go with your first reaction! \texttt{\{Inst.\}} \\

\end{longtable}
\end{center}
\twocolumn

\onecolumn
\begin{center}
\small
\begin{longtable}{p{2cm}p{12.5cm}}
\caption{Examples of prompt variants with different modification types used for the emotion classification task.}
\label{tab:prompt-emo} \\

\toprule
\textbf{Type} & \textbf{Prompt} \\
\midrule
\endfirsthead

\multicolumn{2}{c}{{\tablename\ \thetable{} -- continued from previous page}} \\
\toprule
\textbf{Type} & \textbf{Prompt} \\
\midrule
\endhead

\midrule
\multicolumn{2}{r}{{Continued on next page}} \\
\bottomrule
\endfoot

\bottomrule
\endlastfoot

Base & How would you describe the emotion you infer from the following text using one of the labels `pride', `sadness', `boredom', `anger', `joy', `surprise', `fear', `guilt', `disgust', `trust', `relief', `shame', or `no-emotion'? \texttt{\{Inst.\}} \\
Imper & Describe the emotion you infer from the following text using one of the labels `pride', `sadness', `boredom', `anger', `joy', `surprise', `fear', `guilt', `disgust', `trust', `relief', `shame', or `no-emotion'. \texttt{\{Inst.\}} \\
Imper-pls & Please describe the emotion you infer from the following text using one of the labels `pride', `sadness', `boredom', `anger', `joy', `surprise', `fear', `guilt', `disgust', `trust', `relief', `shame', or `no-emotion'. \texttt{\{Inst.\}} \\
LabelOrd-shuff-1 & How would you describe the emotion you infer from the following text using one of the labels `no-emotion', `joy', `surprise', `pride', `trust', `relief', `sadness', `boredom', `anger', `fear', `guilt', `disgust', or `shame'? \texttt{\{Inst.\}} \\
LabelOrd-shuff-2 & How would you describe the emotion you infer from the following text using one of the labels `joy', `surprise', `pride', `trust', `relief', `no-emotion', `sadness', `boredom', `anger', `fear', `guilt', `disgust', or `shame'? \texttt{\{Inst.\}} \\
LabelOrd-shuff-3 & How would you describe the emotion you infer from the following text using one of the labels `sadness', `boredom', `anger', `fear', `guilt', `disgust', `shame', `no-emotion', `joy', `surprise', `pride', `trust', or `relief'? \texttt{\{Inst.\}} \\
LabelOrd-shuff-4 & How would you describe the emotion you infer from the following text using one of the labels `surprise', `fear', `shame', `boredom', `joy', `guilt', `sadness', `trust', `anger', `relief', `pride', `no-emotion', or `disgust'? \texttt{\{Inst.\}} \\
LabelOrd-shuff-5 & How would you describe the emotion you infer from the following text using one of the labels `shame', `sadness', `trust', `surprise', `relief', `guilt', `joy', `fear', `disgust', `pride', `no-emotion', `boredom', or `anger'? \texttt{\{Inst.\}} \\
LabelOrd-shuff-6 & How would you describe the emotion you infer from the following text using one of the labels `fear', `shame', `no-emotion', `disgust', `surprise', `pride', `relief', `anger', `boredom', `joy', `trust', `sadness', or `guilt'? \texttt{\{Inst.\}} \\
LabelPos-start & Using one of the labels `pride', `sadness', `boredom', `anger', `joy', `surprise', `fear', `guilt', `disgust', `trust', `relief', `shame', or `no-emotion', how would you describe the emotion you infer from the following text? \texttt{\{Inst.\}} \\
Syns-verb-1 & How would you characterize the emotion you infer from the following text using one of the labels `pride', `sadness', `boredom', `anger', `joy', `surprise', `fear', `guilt', `disgust', `trust', `relief', `shame', or `no-emotion'? \texttt{\{Inst.\}} \\
Syns-verb-2 & How would you identify the emotion you infer from the following text using one of the labels `pride', `sadness', `boredom', `anger', `joy', `surprise', `fear', `guilt', `disgust', `trust', `relief', `shame', or `no-emotion'? \texttt{\{Inst.\}} \\
Syns-verb-3 & How would you categorize the emotion you infer from the following text using one of the labels `pride', `sadness', `boredom', `anger', `joy', `surprise', `fear', `guilt', `disgust', `trust', `relief', `shame', or `no-emotion'? \texttt{\{Inst.\}} \\
Syns-noun & How would you describe the emotion you infer from the following text using one of the categories `pride', `sadness', `boredom', `anger', `joy', `surprise', `fear', `guilt', `disgust', `trust', `relief', `shame', or `no-emotion'? \texttt{\{Inst.\}} \\
Syns-prep-1 & How would you describe the emotion you infer from the following text given the labels `pride', `sadness', `boredom', `anger', `joy', `surprise', `fear', `guilt', `disgust', `trust', `relief', `shame', or `no-emotion'? \texttt{\{Inst.\}} \\
Syns-prep-2 & How would you describe the emotion you infer from the following text according to the labels `pride', `sadness', `boredom', `anger', `joy', `surprise', `fear', `guilt', `disgust', `trust', `relief', `shame', or `no-emotion'? \texttt{\{Inst.\}} \\
Syns-co-1 & How would you describe the emotion you infer from the following text using one of the labels `pride', `sadness', `boredom', `anger', `joy', `surprise', `fear', `guilt', `disgust', `trust', `relief', `shame', and `no-emotion'? \texttt{\{Inst.\}} \\
Syns-co-2 & How would you describe the emotion you infer from the following text using one of the labels `pride', `sadness', `boredom', `anger', `joy', `surprise', `fear', `guilt', `disgust', `trust', `relief', `shame', `no-emotion'? \texttt{\{Inst.\}} \\
Typo-task-1 & How would you describe the emoton you infer from the following text using one of the labels `pride', `sadness', `boredom', `anger', `joy', `surprise', `fear', `guilt', `disgust', `trust', `relief', `shame', or `no-emotion'? \texttt{\{Inst.\}} \\
Typo-task-2 & How would you describ the emotion you inferr from the following text using one of the labels `pride', `sadness', `boredom', `anger', `joy', `surprise', `fear', `guilt', `disgust', `trust', `relief', `shame', or `no-emotion'? \texttt{\{Inst.\}} \\
Typo-task-3 & How would you describe the emotion you infer from the following text using one of the lables `pride', `sadness', `boredom', `anger', `joy', `surprise', `fear', `guilt', `disgust', `trust', `relief', `shame', or `no-emotion'? \texttt{\{Inst.\}} \\
Typo-label-1 & How would you describe the emotion you infer from the following text using one of the labels `prde', `sadnes', `bordom', `anger', `joy', `suprise', `fear', `guilt', `disgst', `turst', `relif', `shame', or `no-emotin'? \texttt{\{Inst.\}} \\
Typo-label-2 & How would you describe the emotion you infer from the following text using one of the labels `pride', `sadness', `boredom', `angre', `joy', `surprise', `feer', `guillt', `disgust', `trust', `relief', `shaem', or `no-emotion'? \texttt{\{Inst.\}} \\

Typo-label-3 & How would you describe the emotion you infer from the following text using one of the labels `pried', `sadnes', `bordom', `angr', `joy', `suprise', `feer', `guillt', `disgst', `turst', `relif', `shaem', or `noemotion'? \texttt{\{Inst.\}} \\
Cap-task & How would you describe the EMOTION you infer from the following TEXT using one of the labels `pride', `sadness', `boredom', `anger', `joy', `surprise', `fear', `guilt', `disgust', `trust', `relief', `shame', or `no-emotion'? \texttt{\{Inst.\}} \\
Cap-label & How would you describe the emotion you infer from the following text using one of the labels `PRIDE', `SADNESS', `BOREDOM', `ANGER', `JOY', `SURPRISE', `FEAR', `GUILT', `DISGUST', `TRUST', `RELIEF', `SHAME', or `NO-EMOTION'? \texttt{\{Inst.\}} \\
PM-1 & How would you describe the emotion you infer from the following text using one of the labels pride, sadness, boredom, anger, joy, surprise, fear, guilt, disgust, trust, relief, shame, or no-emotion? \texttt{\{Inst.\}} \\
PM-2 & How would you describe the emotion you infer from the following text using one of the labels: `pride', `sadness', `boredom', `anger', `joy', `surprise', `fear', `guilt', `disgust', `trust', `relief', `shame', or `no-emotion'? \texttt{\{Inst.\}} \\
PM-3 & How would you describe the emotion you infer from the following text using one of the labels `pride'; `sadness'; `boredom'; `anger'; `joy'; `surprise'; `fear'; `guilt'; `disgust'; `trust'; `relief'; `shame'; or `no-emotion'? \texttt{\{Inst.\}} \\
AltLab-keep-1 & How would you describe the emotion you infer from the following text using one of the labels `self-satisfaction', `sorrow', `disinterest', `rage', `happiness', `amazement', `panic', `remorse', `revulsion', `belief', `comfort', `embarrassment', or `neutral'? \texttt{\{Inst.\}} \\
AltLab-keep-2 & How would you describe the emotion you infer from the following text using one of the labels `self-respect', `grief', `monotony', `fury', `delight', `astonishment', `terror', `regret', `aversion', `confidence', `release', `humiliation', or `emotionless'? \texttt{\{Inst.\}} \\
AltLab-int & How would you describe the emotion you infer from the following text using one of the labels `pride', `sadness', `boredom', `anger', `joy', `surprise', `fear', `guilt', `disgust', `trust', `relief', or `shame'? \texttt{\{Inst.\}} \\
Conf & How would you describe the emotion you infer from the following text using one of the labels `pride', `sadness', `boredom', `anger', `joy', `surprise', `fear', `guilt', `disgust', `trust', `relief', `shame', or `no-emotion'? Provide the label along with a confidence score (1-5) of your selection, and omit any justification. \\
Exp & How would you describe the emotion you infer from the following text using one of the labels `pride', `sadness', `boredom', `anger', `joy', `surprise', `fear', `guilt', `disgust', `trust', `relief', `shame', or `no-emotion'? Provide the label along with a justification for your selection. \\
Emo-1 & How would you describe the emotion you infer from the following text using one of the labels `pride', `sadness', `boredom', `anger', `joy', `surprise', `fear', `guilt', `disgust', `trust', `relief', `shame', or `no-emotion'? Important: Trust your gut reaction! \texttt{\{Inst.\}} \\
Emo-2 & How would you describe the emotion you infer from the following text using one of the labels `pride', `sadness', `boredom', `anger', `joy', `surprise', `fear', `guilt', `disgust', `trust', `relief', `shame', or `no-emotion'? Warning: Some texts may be disturbing. \texttt{\{Inst.\}} \\
Emo-3 & How would you describe the emotion you infer from the following text using one of the labels `pride', `sadness', `boredom', `anger', `joy', `surprise', `fear', `guilt', `disgust', `trust', `relief', `shame', or `no-emotion'? Please be careful with your selection. \texttt{\{Inst.\}} \\
Emo-4 & How would you describe the emotion you infer from the following text using one of the labels `pride', `sadness', `boredom', `anger', `joy', `surprise', `fear', `guilt', `disgust', `trust', `relief', `shame', or `no-emotion'? Important: Choose quickly and go with your first reaction! \texttt{\{Inst.\}} \\
Def-task & How would you describe the emotion you infer from the following text using one of the labels `pride', `sadness', `boredom', `anger', `joy', `surprise', `fear', `guilt', `disgust', `trust', `relief', `shame', or `no-emotion'? Your choice should reflect one primary emotion you infer the author or speaker is expressing, based on the tone, context, and implied sentiment of the text. \texttt{\{Inst.\}} \\
Def-label-1 & How would you describe the emotion you infer from the following text using one of these labels?
        \newline`pride': A sense of self-respect or accomplishment.
        \newline`sadness': Emotional pain, sorrow, or unhappiness.
        \newline`boredom': Lack of interest or engagement.
        \newline`anger': Strong displeasure or hostility.
        \newline`joy': Intense happiness or delight.
        \newline`surprise': Unexpectedness or astonishment.
        \newline`fear': Anxiety or distress caused by threat.
        \newline`guilt': Remorse over wrongdoing.
        \newline`disgust': Revulsion or strong disapproval.
        \newline`trust': Confidence in reliability or honesty.
        \newline`relief': Reassurance after distress.
        \newline`shame': Humiliation or embarrassment.
        \newline`no-emotion': Neutral tone; absence of discernible feeling.
        \newline\texttt{\{Inst.\}} \\

Def-label-2 & How would you describe the emotion you infer from the following text using one of these labels?
        \newline`pride': A feeling of self-respect, achievement, or personal worth.
        \newline`sadness': A state of unhappiness, grief, or emotional pain.
        \newline`boredom': A lack of interest or engagement; feeling unstimulated.
        \newline`anger': A strong feeling of displeasure or hostility.
        \newline`joy': A feeling of great happiness or delight.
        \newline`surprise': A reaction to something unexpected or sudden.
        \newline`fear': An emotional response to threat or danger, real or perceived.
        \newline`guilt': A feeling of responsibility or remorse for a wrongdoing.
        \newline`disgust': A sense of revulsion or profound disapproval.
        \newline`trust': Confidence or belief in the reliability or integrity of someone/something.
        \newline`relief': A feeling of reassurance or alleviation of distress.
        \newline`shame': A painful feeling of humiliation or embarrassment due to perceived wrongdoing.
        \newline`no-emotion': Used when the text is emotionally neutral or lacks a discernible emotional tone.
        \newline\texttt{\{Inst.\}} \\
Def-both & How would you describe the emotion you infer from the following text using one of these labels? Your choice should reflect one primary emotion you infer the author or speaker is expressing, based on the tone, context, and implied sentiment of the text.
        \newline`pride': A feeling of self-respect, achievement, or personal worth.
        \newline`sadness': A state of unhappiness, grief, or emotional pain.
        \newline`boredom': A lack of interest or engagement; feeling unstimulated.
        \newline`anger': A strong feeling of displeasure or hostility.
        \newline`joy': A feeling of great happiness or delight.
        \newline`surprise': A reaction to something unexpected or sudden.
        \newline`fear': An emotional response to threat or danger, real or perceived.
        \newline`guilt': A feeling of responsibility or remorse for a wrongdoing.
        \newline`disgust': A sense of revulsion or profound disapproval.
        \newline`trust': Confidence or belief in the reliability or integrity of someone/something.
        \newline`relief': A feeling of reassurance or alleviation of distress.
        \newline`shame': A painful feeling of humiliation or embarrassment due to perceived wrongdoing.
        \newline`no-emotion': Used when the text is emotionally neutral or lacks a discernible emotional tone.
        \newline\texttt{\{Inst.\}} \\

\end{longtable}
\end{center}
\twocolumn


\end{document}